\pdfoutput=1
\documentclass[11pt]{article}

\usepackage[final]{acl}

\usepackage{iftex}
\newif\ifcjk \newif\ifarab
  \usepackage[T1]{fontenc}
  \usepackage[utf8]{inputenc}
  \usepackage{times}
  \usepackage{inconsolata}
  \cjkfalse \arabfalse
\newcommand{\zh}[2]{\ifcjk{\cjkfont #1} #2\else #2\fi}

\usepackage{latexsym}
\usepackage{microtype}
\usepackage{graphicx}
\usepackage{booktabs}
\usepackage{makecell}
\usepackage{amsmath}
\usepackage{amssymb}
\usepackage{xspace}
\usepackage{float}
\usepackage{longtable}
\usepackage{placeins}
\usepackage{enumitem}

\newcommand{\method}{SAE\mbox{-}IDF\xspace}

\newcommand{\pcone}{Dense\textsubscript{$-$PC1}\xspace}
\newcommand{\R}{\mathbb{R}}
\newcommand{\idf}{\mathrm{idf}}
\newcommand{\df}{\mathrm{df}}
\DeclareMathOperator*{\argmax}{arg\,max}

\title{Sparse Autoencoders Are a Shared Coordinate System\\for Symbolic Matching}

\author{Xiaocong Yang \quad Xuanyu Meng \quad Yuji Zhang \quad Jiawei Han \quad ChengXiang Zhai \\
  University of Illinois Urbana-Champaign \\
  \texttt{\{xy51, xuanyum2, yujiz, hanj, czhai\}@illinois.edu}}

\begin{document}
\maketitle

\begin{abstract}
Large language models (LLMs) hold broad knowledge in one set of weights, but lack interpretability and trustworthy reasoning, especially in high-stakes use cases. A neuro-symbolic model, with its interpretable architecture and System~2 reasoning capacity, is a promising alternative if built at the scale of LLMs; however, its symbolic backbone is usually constructed with human effort, so existing neuro-symbolic models are bottlenecked by the scope of the symbol component and usually domain-specific. A data-driven method could instead build a larger symbolic system by aligning and merging existing ones. We target its first, foundational step: matching units across these heterogeneous systems. Our method, \method, encodes a symbol in its native context into a sparse code with a sparse autoencoder (SAE), re-weights each dimension by its rarity and matches across systems.
On 12 tasks across 6 kinds of symbolic system, \method outperforms the same LLM's dense baseline on all tasks with Gemma-2-9B and all but one with Llama-3.1-8B, by up to 45 points, consistently across SAE layers and configurations.
\method is also interpretable: every dimension of the sparse code has a human-readable meaning, so each match decomposes into named concepts (features).
Our work paves the way to a universal symbolic backbone.
\end{abstract}

\section{Introduction}
\label{sec:intro}

\begin{figure*}[t]
  \centering
  \includegraphics[width=\textwidth]{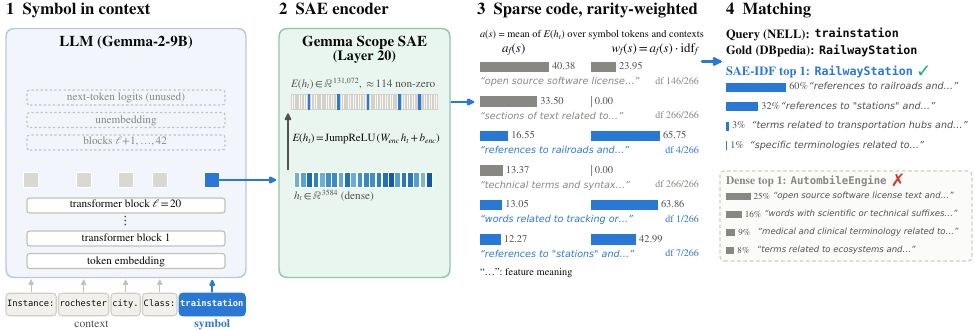}
  \caption{\method on the NELL class \texttt{trainstation} (Common-KG NELL $\to$ DBpedia). (1)~A native view---context first, symbol last---goes through the LLM; the residual stream $h_t$ is read at the symbol's tokens. (2)~The SAE encoder $E$ maps each state to a sparse code; codes are averaged into $a(s)$. (3)~Its largest coordinates, with explanations and document frequencies: idf damps features that fire on many symbols (grey). (4)~Each bar is a feature's share of the cosine with the query; the mark says whether it is the gold: 93\% of the match with the gold \texttt{RailwayStation} comes from two features for railroads and stations, whereas dense ranks the gold fifth and \texttt{AutombileEngine} (sic) first, which shares only ubiquitous or unrelated features and ranks fourth under \method.}
  \label{fig:pipeline}
\end{figure*}

Modern large language models (LLMs) hold broad knowledge in a single set of weights and are widely seen as a route to general-purpose intelligence. However, the lack of interpretability of a model's internal working mechanism \citep{zhao2024explainability}, and the controversy on whether it has genuine reasoning capability rather than pattern matching \citep{mirzadeh2025gsmsymbolic, wu2024reasoning}, make that route questionable \citep{dawid2024energy}, especially in high-stakes scenarios such as healthcare. Neuro-symbolic models \citep{garcez2023neurosymbolic, kautz2022third}, instead, address the two defects of LLMs architecturally: a symbolic component exposes intermediate states that a person can read and intervene on \citep{yang2026generative}, and supplies the slow, deliberate, and usually rule-governed reasoning capability that cognitive theory calls System~2 \citep{kahneman2011thinking, booch2021thinking, goyal2022inductive}. 

Yet a major challenge that blocks neuro-symbolic models is scalability. Unlike LLMs whose knowledge is obtained via self-supervised learning on an enormous corpus, which benefits from modern computing hardware, symbolic systems depend on knowledge and careful handcraft from human experts. Consequently, most existing symbolic systems and neuro-symbolic models are confined to one domain \citep{mathlib2020, ashburner2000gene, silver2016alphago, segler2018planning, lample2020deep}.

Nonetheless, much symbolic knowledge does already exist, and AI now adds to it, e.g., Lean proofs on open problems \citep{openai2026math}. But it is spread across those numerous domain-specific systems that each describe it \textit{in their own schema}: a math problem is a Lean~4 \texttt{theorem} in one proof assistant and a string of \texttt{\$e} hypotheses in Metamath, and the same place is \texttt{TrainStation} in one knowledge graph and \texttt{RailwayStation} in another. Indeed, aligning such heterogeneous systems is a classic problem in several communities---ontology matching \citep{euzenat2013ontology}, schema matching \citep{rahm2001survey}, entity alignment \citep{sun2020benchmarking}, cross-language code retrieval \citep{zhu2022xlcost}---but each solves it for its own kind of system. We use one term, \emph{symbolic matching}, for all of them, and a general approach to it is the first step towards the universal symbolic backbone that a general-purpose neuro-symbolic model needs, reusing the human effort in existing systems instead of starting over like Cyc \citep{lenat1995cyc}.

One might think a pretrained LLM already supplies a general tool for symbolic matching: trained on a large general corpus, it reads almost any notation and contextualizes each symbol through attention \citep{attention}.
Its hidden states are a natural candidate, but they are known to be anisotropic, dominated by directions shared by all inputs \citep{ethayarajh2019contextual, mu2018allbutthetop}, and dense text embeddings tend to favor superficial linguistic similarity over structural correspondence \citep{yang2024analogy}.
Neither defect is easy to fix: each dense dimension mixes many concepts in superposition \citep{elhage2022toy}, so it can be neither rarity-weighted nor read to explain a match, and post-hoc fixes such as whitening trail \method on most tasks (Appendix~\ref{app:controls}).

Sparse autoencoders (SAEs) address these defects. They write an LLM's hidden state as a sparse sum of interpretable \textit{features} from a large dictionary \citep{bricken2023monosemanticity, cunningham2024sparse, templeton2024scaling}. Trained on a large corpus, the dictionary is a \textit{shared coordinate system} for any token, including symbols: each feature is a basis direction and its weight (the \textit{code}) the coordinate. Here, the directions shared by every dense vector become a few named, ubiquitous features that can be identified and downweighted, as well as lexical features, so each symbol's contextual meaning contributes most to the matching.

Upon such merits, we build \method, a general approach for symbolic matching. An LLM reads each symbol in its \emph{native context}; the SAE encodes it as a sparse code, in effect a bag of named concepts, so text retrieval's classic inverse document frequency (IDF; \citealp{sparckjones1972statistical}) applies: features are weighted by rarity and candidates ranked by the cosine of weighted codes (Figure~\ref{fig:pipeline}), much as DNA sequence search finds relatedness in a few high-scoring local matches, each scored against chance \citep{altschul1990blast}, not in stretches every genome shares. Needing no aligned pairs, \method matches any systems an LLM can read.

Across 12 tasks on 6 kinds of symbolic system, \method beats the same model's dense hidden state on every task with Gemma-2-9B and on all but one with Llama-3.1-8B, and it is the best training-free method on most tasks.

\paragraph{Our main contributions are as follows:}

\begin{itemize}[leftmargin=*, nosep, topsep=1pt]
  \item \method, a training-free, general and interpretable approach to symbolic matching that significantly outperforms baselines (Sections~\ref{sec:method},~\ref{sec:experiments}).
  \item Ablations showing that both ingredients matter and the gain is robust (Sections~\ref{sec:ablation},~\ref{sec:sweeps}).
  \item A feature-level interpretability analysis with rule-selected case studies, a column-type split and merge tests (Sections~\ref{sec:interp},~\ref{sec:discussion}).
\end{itemize}

\section{\method}
\label{sec:method}

Symbolic matching takes a query symbol $s$ of a source system $A$ and ranks a candidate set $C(s)$ from a target system $B$ so that the counterpart of $s$ comes first.
Let $h_t\in\R^d$ be the residual stream of a causal language model at layer $\ell$ and token position $t$ when it reads a text, and let $E:\R^d\to\R^F_{\ge 0}$ be the encoder of a sparse autoencoder with $F$ features trained on that layer; for a JumpReLU SAE, $E(h)=\mathrm{JumpReLU}(W_{\mathrm{enc}}h+b_{\mathrm{enc}})$ with $\mathrm{JumpReLU}(z)_f=z_f\,\mathbf{1}[z_f>\theta_f]$ and a learned threshold $\theta_f$ per feature \citep{rajamanoharan2024jumping}.
\method represents every symbol by a sparse vector (code) in $\R^F$ with the following pipeline before matching with cosine similarity.

\paragraph{Native contexts.}
Every symbol $s$ comes with pieces of information recorded in its own system, which we call its \emph{views} $v_1(s),\dots,v_m(s)$: the instances of a knowledge-graph class, the axioms an ontology term occurs in, or the sampled cell values of a table column.
Each view is turned into part of input to the LLM, consisting of a \emph{context} $c_j(s)$, followed by the symbol itself (Figure~\ref{fig:pipeline}).
The LLM reads each input in a forward pass to the target layer $\ell$, whose residual-stream states $\{h_t\}$ are handed to the SAE of that layer in the next step.
Table~\ref{tab:benchmarks} lists the views and templates of every dataset.

\paragraph{Sparse code.}
Of each input only the states at the symbol's own token positions $T_j(s)$ are used, as the information from context has been extracted and added to those tokens via self-attention. When the whole text is the symbol---a program, a formal statement, or a table column written as its name and sampled values---$T_j(s)$ is every position.
Each kept token is SAE-encoded first and the codes are averaged afterwards, over tokens and then views:
\begin{equation}
  a(s) \;=\; \frac{1}{m}\sum_{j=1}^{m}\frac{1}{|T_j(s)|}\sum_{t\in T_j(s)} E(h_t)\;\in\R^{F}.
  \label{eq:code}
\end{equation}

where $a(s)$, the \emph{sparse code vector}, has non-negative and mostly zero elements. 

\paragraph{Rarity weighting.}
Over a reference collection $D$ of $N$ symbols, let $\df(f)=|\{u\in D : a_f(u)>0\}|$ be the number of symbols on which feature $f$ fires.
Then each dimension of $a(s)$ is multiplied by its feature's inverse document frequency,
\begin{equation}
  w(s) = a(s)\odot\idf\in\R^F,\; \idf_f=\Big[\log\tfrac{N}{1+\df(f)}\Big]_+ ,
  \label{eq:idf}
\end{equation}
so a feature firing on every symbol gets weight zero and one firing on few symbols is amplified.

\begin{table*}[t]
  \centering\footnotesize
  \setlength{\tabcolsep}{3pt}
  \begin{tabular}{@{}llrrrrrrrrr@{}}
\toprule
 & & & \multicolumn{4}{c}{Gemma-2-9B (layer 20)} & \multicolumn{4}{c}{Llama-3.1-8B (layer 15)} \\
\cmidrule(lr){4-7}\cmidrule(lr){8-11}
Task & Metric & Strings & Dense & \makecell{Dense,\\PC1 rm.} & \textbf{SAE-IDF} & \textbf{$\boldsymbol{\Delta}$ dense} & Dense & \makecell{Dense,\\PC1 rm.} & \textbf{SAE-IDF} & \textbf{$\boldsymbol{\Delta}$ dense} \\
\midrule
\multicolumn{11}{@{}l}{\textit{Formal mathematics (miniF2F)}} \\
\quad Lean 4 $\to$ Isabelle & MRR & 0.825 & 0.912 & 0.926 & \textbf{0.961} & \textbf{+0.05} & 0.942 & 0.944 & \textbf{0.976} & \textbf{+0.03} \\
\quad Lean 4 $\to$ Metamath & MRR & 0.615 & 0.210 & 0.151 & \textbf{0.661} & \textbf{+0.45} & 0.646 & 0.637 & \textbf{0.787} & \textbf{+0.14} \\
\quad Lean 4 $\to$ HOL Light & MRR & 0.741 & 0.539 & 0.531 & \textbf{0.898} & \textbf{+0.36} & 0.603 & 0.699 & \textbf{0.906} & \textbf{+0.30} \\
\multicolumn{11}{@{}l}{\textit{Code (XLCoST, program level)}} \\
\quad 7 languages, mean & MRR$^\dagger$ & \textbf{0.468} & 0.454 & 0.452 & 0.464 & \textbf{+0.01} & 0.451 & 0.447 & 0.442 & $\boldsymbol{-}$0.01 \\
\quad 7 languages, mean & P@6$^\dagger$ & \textbf{0.745} & 0.680 & 0.678 & 0.732 & \textbf{+0.05} & 0.641 & 0.648 & 0.657 & \textbf{+0.02} \\
\multicolumn{11}{@{}l}{\textit{Table schemas (Valentine)}} \\
\quad source $\to$ target columns & F1 & -- & 0.533 & 0.534 & \textbf{0.614} & \textbf{+0.08} & 0.588 & 0.579 & \textbf{0.625} & \textbf{+0.04} \\
\quad source $\to$ target columns & MRR & -- & 0.766 & 0.761 & \textbf{0.856} & \textbf{+0.09} & 0.786 & 0.811 & \textbf{0.862} & \textbf{+0.08} \\
\multicolumn{11}{@{}l}{\textit{Knowledge graphs (OAEI Common-KG)}} \\
\quad NELL $\to$ DBpedia & MRR & 0.936 & 0.931 & 0.957 & \textbf{0.976} & \textbf{+0.04} & 0.949 & \textbf{0.960} & 0.958 & \textbf{+0.01} \\
\quad YAGO $\to$ Wikidata & MRR & 0.899 & 0.860 & 0.932 & \textbf{0.962} & \textbf{+0.10} & 0.889 & \textbf{0.944} & \textbf{0.944} & \textbf{+0.06} \\
\multicolumn{11}{@{}l}{\textit{Cross-lingual ontologies (MultiFarm)}} \\
\quad Chinese $\to$ English & MRR & 0.230 & 0.650 & \textbf{0.760} & 0.754 & \textbf{+0.10} & 0.659 & 0.779 & \textbf{0.787} & \textbf{+0.13} \\
\quad Russian $\to$ English & MRR & 0.234 & 0.644 & 0.751 & \textbf{0.765} & \textbf{+0.12} & 0.661 & \textbf{0.751} & 0.726 & \textbf{+0.06} \\
\quad Arabic $\to$ English & MRR & 0.238 & 0.595 & 0.702 & \textbf{0.722} & \textbf{+0.13} & 0.578 & 0.669 & \textbf{0.672} & \textbf{+0.09} \\
\multicolumn{11}{@{}l}{\textit{Biomedical ontologies (Bio-ML, Anatomy)}} \\
\quad NCIT $\to$ DOID & MRR & 0.779 & 0.701 & 0.732 & \textbf{0.817} & \textbf{+0.12} & 0.704 & 0.710 & \textbf{0.779} & \textbf{+0.08} \\
\quad mouse $\to$ human & MRR & 0.734 & 0.812 & 0.864 & \textbf{0.886} & \textbf{+0.07} & 0.800 & \textbf{0.871} & 0.844 & \textbf{+0.04} \\
\bottomrule
\end{tabular}

  \caption{Main results with native context for two model families, each with a 131k-feature SAE (Gemma Scope, Llama Scope). Bold: best per row within each model, Strings in both; ties go to the model. $\Delta$ dense: \method minus Dense of the same model. $^\dagger$XLCoST official metrics averaged over the seven query languages (Appendix~\ref{app:xlcost}); Strings is word TF-IDF there, and Valentine has no comparable string baseline. Trained retrievers: Appendix~\ref{app:embed}.}
  \label{tab:main}
\end{table*}

\paragraph{Matching.}
Every symbol is encoded into a sparse code with the pipeline above. Given a query symbol $s$ and a pool of candidate symbols $C(s)$, candidates are ranked by the cosine of the weighted vectors, and by default the top-1 candidate is taken as the predicted matched symbol: 

\begin{equation}
    \hat t(s)=\argmax_{t\in C(s)}\cos\big(w(s),w(t)\big)
\end{equation}

The cosine decomposes exactly over coordinates of all features: $\cos(w(s),w(t))=\sum_f c_f$ with $c_f=w_f(s)\,w_f(t)/(\|w(s)\|\,\|w(t)\|)$, where the contribution of each interpretable feature to the similarity score is available.

\section{Experiments}
\label{sec:experiments}

\subsection{Datasets}
\label{sec:datasets}

Six different kinds of symbolic systems were chosen to test the universal applicability of \method, where every run uses the full test data and the benchmark's own metric code where one exists.
miniF2F \citep{zheng2022minif2f}: from the Lean~4 port \citep{yang2023minif2flean4} we retrieve each problem's statement in another prover among all statements of that prover, with names, proofs, comments and labels stripped so that nothing but the mathematics identifies a problem.
XLCoST \citep{zhu2022xlcost}: the official protocol and evaluator (the pool contains the query's own row, a free first hit); MRR and Precision@6 averaged over the seven query languages.
Valentine \citep{koutras2021valentine}: all 551 joinable and unionable pairs, scored with Valentine's F1 (one-to-one assignment) and MRR.
Common-KG \citep{fallatah2020gold}: 129 NELL--DBpedia reference pairs and 304 YAGO--Wikidata pairs; classes without instances match on their label alone.
MultiFarm \citep{meilicke2012multifarm}: foreign-language terms against the English terms of a \emph{different} conference ontology, so that neither language nor ontology is shared.
Bio-ML \citep{he2022bioml} and OAEI Anatomy \citep{bodenreider2005mice}: biomedical ontologies, NCIT $\to$ DOID on the Bio-ML validation split with its 100-candidate ranking protocol, and mouse $\to$ human anatomy against all 3,298 human classes; about half of their gold pairs have identical labels.
The metric is MRR unless stated; Appendix~\ref{app:benchmarks} gives view formats.
PubChem SMILES / IUPAC / InChI \citep{kim2023pubchem}, notations the model cannot read, are in Appendix~\ref{app:tasks}.

\subsection{Setup and baselines}
\label{sec:baselines}

In the main experiments, we choose two model families: Gemma-2-9B \citep{gemmateam2024gemma2} at $\ell=20$ with the Gemma Scope JumpReLU SAE ($F=131{,}072$, average $L_0$ of 114; \citealp{lieberum2024gemmascope}), and Llama-3.1-8B \citep{grattafiori2024llama3} at $\ell=15$ with the Llama Scope SAE of the same width \citep{he2024llamascope}; other layers, widths, sparsities and model sizes are swept in Section~\ref{sec:sweeps}.
The reference collection $D$ for idf is the candidate pool for miniF2F, XLCoST and Valentine and all symbols of both systems for the ontology and knowledge-graph tasks (Appendix~\ref{app:benchmarks}).

All baselines use exactly the same views and token positions.
\emph{Dense} is the hidden state averaged over symbol tokens and views, similar to the form in Eq.~\ref{eq:code} except for the use SAE.
\emph{\pcone} additionally subtracts the mean of the dense vectors over $D$ and removes their projection onto the first principal direction $u_1$ of $D$, $(I-u_1u_1^\top)(d-\mu)$, the common-direction removal of \citet{mu2018allbutthetop} and \citet{arora2017simple}.
\emph{SAE-unweighted} is the unweighted code $a(s)$, which isolates the sparse basis from the weighting.
\emph{Strings} is the best of word TF-IDF \citep{sparckjones1972statistical}, character 3--5-gram TF-IDF and BM25 \citep{robertson2009probabilistic} over the surface strings of the same views, chosen per task in the strings' favour (Appendix~\ref{app:benchmarks}); TF-IDF over strings separates the SAE features from the weighting.
\emph{Retrievers}: BGE-M3 \citep{chen2024bge} and Qwen3-Embedding-8B \citep{zhang2025qwen3embedding}, embedding models specially trained for retrieval, on the same view texts with the same controls (Appendix~\ref{app:embed}); every similarity is a cosine. They are a reference rather than a like-for-like baseline: a retriever is trained on paired data to emit one similarity, whereas what is under test here is the untrained feature space of a general model.
We compare with no task-specific matcher (such as BERTMap \citep{he2022bertmap} and LogMap \citep{jimenezruiz2011logmap}), as each is built, tuned and usually trained for one kind of system and does not run on the others and cannot meet the requirement of symbolic matching in general.

\subsection{Main results}
\label{sec:results}

Table~\ref{tab:main} gives the main results (plotted in Figure~\ref{fig:main}, Appendix~\ref{app:tasks}).
With Gemma-2-9B, \method beats the dense hidden state of the same model on every task, by up to +0.45 MRR (Lean~4 $\to$ Metamath), and is ahead of \pcone and strings everywhere except XLCoST (word TF-IDF marginally ahead) and MultiFarm Chinese (\pcone by 0.006).
A paired bootstrap puts every gain over dense at $p<0.004$, and the gain over \pcone at $p<0.01$ on 14 of 18 comparisons (Appendix~\ref{app:significance}).
With Llama-3.1-8B the margins are smaller: \method beats dense on 13 of 14 rows, all but XLCoST MRR, and is ahead of or level with \pcone on 10 of 14, trailing on the other four: XLCoST MRR, NELL $\to$ DBpedia, MultiFarm Russian and Anatomy.

The magnitude of the gain depends on how similar the two surface notations are.
Isabelle statements resemble Lean~4 and the gain is small (dense 0.912, \method 0.961); Metamath and HOL Light read very differently, and dense collapses to 0.210 and 0.539 where \method keeps 0.661 and 0.898.
On MultiFarm, where source and target share neither script nor ontology, strings are near useless (0.23, MRR) and \method gains +0.10 to +0.13 over dense.
Conversely, XLCoST programs share identifiers across languages, so word TF-IDF is strong (0.468) and \method gains only +0.01 over dense.

\begin{figure*}[t]
  \centering
  \includegraphics[width=\textwidth]{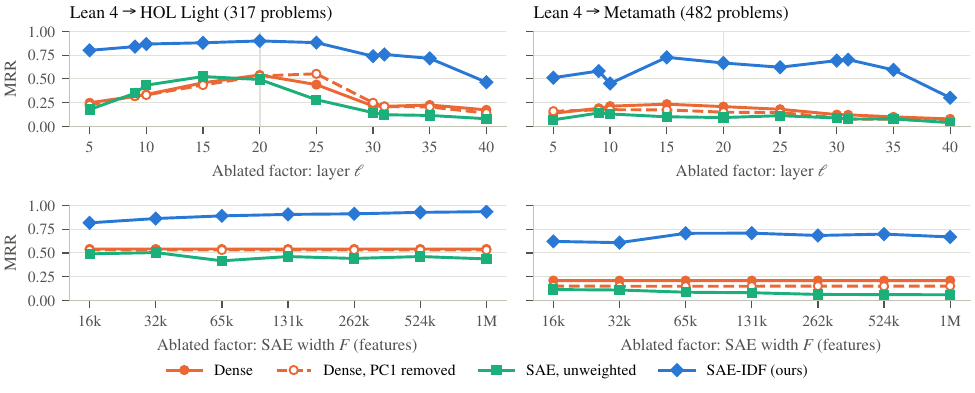}
  \caption{Ablating the layer (top; 131k-feature SAE at each layer, grey line: the main-table layer) and the SAE width at matched sparsity (bottom; layer 20, average $L_0$ 48--58) with Gemma-2-9B on miniF2F Lean~4 $\to$ HOL Light and $\to$ Metamath. \method beats every baseline at every layer and width; the unweighted code tracks or trails dense.}
  \label{fig:ablation}
\end{figure*}

\begin{table}[t]
  \centering\fontsize{8}{9.5}\selectfont
  \setlength{\tabcolsep}{2pt}\renewcommand{\quad}{\hspace{0.5em}}
  \begin{tabular}{@{}llrrr@{}}
\toprule
Task & Metric & \makecell{Random\\+\,idf} & \makecell{SAE\\unweighted} & \textbf{SAE-IDF} \\
\midrule
\multicolumn{5}{@{}l}{\textit{Formal mathematics (miniF2F)}} \\
\quad Lean 4 $\to$ Isabelle & MRR & 0.865 & 0.882 & \textbf{0.961} \\
\quad Lean 4 $\to$ Metamath & MRR & 0.146 & 0.095 & \textbf{0.661} \\
\quad Lean 4 $\to$ HOL Light & MRR & 0.349 & 0.493 & \textbf{0.898} \\
\multicolumn{5}{@{}l}{\textit{Code (XLCoST, program level)}} \\
\quad 7 languages, mean & MRR$^\dagger$ & 0.417 & 0.449 & \textbf{0.464} \\
\quad 7 languages, mean & P@6$^\dagger$ & 0.627 & 0.673 & \textbf{0.732} \\
\multicolumn{5}{@{}l}{\textit{Table schemas (Valentine)}} \\
\quad source $\to$ target columns & F1 & 0.482 & 0.458 & \textbf{0.614} \\
\quad source $\to$ target columns & MRR & 0.647 & 0.685 & \textbf{0.856} \\
\multicolumn{5}{@{}l}{\textit{Knowledge graphs (OAEI Common-KG)}} \\
\quad NELL $\to$ DBpedia & MRR & 0.901 & 0.938 & \textbf{0.976} \\
\quad YAGO $\to$ Wikidata & MRR & 0.797 & 0.872 & \textbf{0.962} \\
\multicolumn{5}{@{}l}{\textit{Cross-lingual ontologies (MultiFarm)}} \\
\quad Chinese $\to$ English & MRR & 0.533 & 0.404 & \textbf{0.754} \\
\quad Russian $\to$ English & MRR & 0.542 & 0.252 & \textbf{0.765} \\
\quad Arabic $\to$ English & MRR & 0.425 & 0.225 & \textbf{0.722} \\
\multicolumn{5}{@{}l}{\textit{Biomedical ontologies (Bio-ML, Anatomy)}} \\
\quad NCIT $\to$ DOID & MRR & 0.692 & 0.696 & \textbf{0.817} \\
\quad mouse $\to$ human & MRR & 0.788 & 0.801 & \textbf{0.886} \\
\bottomrule
\end{tabular}

  \caption{Ablating the two ingredients of \method on the main-table tasks. \emph{Random\,+\,idf}: a random dictionary of the same width and sparsity with the same rarity weighting; \emph{SAE-unweighted}: the learned features without it. Bold: best per row. $^\dagger$As in Table~\ref{tab:main}.}
  \label{tab:ablationmain}
\end{table}

Removing the first principal component from dense hidden states recovers some of the gap on ontology and knowledge-graph terms, whose dense vectors share a large common component, but makes no difference on code, formal statements and tables.
Stronger controls (Appendix~\ref{app:controls})---top-36 PC removal, standardization, whitening, a dense analogue of idf---sharpen the picture.

Even without any training on retrieval objectives, \method outperforms the dedicated multilingual retriever BGE-M3 on 10 of 15 rows (same views, same controls; Appendix~\ref{app:embed}), and against Qwen3-Embedding-8B, a retriever of the same scale as our backbone, it leads on code, YAGO $\to$ Wikidata, MultiFarm Arabic and Bio-ML while trailing on 8 rows, mainly formal mathematics (0.97--0.99 MRR), while it might be attributed to memorization from pretraining data: these datasets have been public under matching names since 2021 \citep{zheng2022minif2f}, in Lean~4 since 2023 \citep{yang2023minif2flean4}, and Qwen3's 36T-token pretraining corpus includes code and STEM data whose contents are not disclosed \citep{yang2025qwen3}. Our backbone and its SAE may have read these statements as well, but they are trained only to predict or reconstruct text, whereas a retriever is further trained on query--document pairs to score each query's counterpart above other candidates \citep{chen2024bge, zhang2025qwen3embedding}, which is exactly how matching is evaluated here.
More importantly, our approach is naturally aligned with the trend of scaling up LLMs and their SAEs: sparse dictionaries now reach tens of millions of features on frontier models \citep{templeton2024scaling, gao2025scaling} and are released for every layer of open models \citep{lieberum2024gemmascope, he2024llamascope}, and \method inherits every such release at no training cost.
The empirical result confirms the benefit of scaling up in symbolic matching tasks: for example, HOL Light MRR rises from 0.87 to 0.94 and Metamath from 0.65 to 0.74 across Gemma-2-2B, 9B and 27B (Section~\ref{sec:sweeps}), whereas a retriever is bounded by its training pairs and must be retrained at each size. A retriever must also be trained on aligned pairs from the domain it will serve, which a newly encountered pair of systems does not have; \method needs none, which is why one method covers all six kinds of system here.

\section{Ablations and Analysis}
\label{sec:analysis}

\subsection{Rarity weighting and the coordinate system}
\label{sec:ablation}

We ablate the two factors of \method, the rarity weighting and the learned coordinate system, one at a time. Without the $idf$ factor, the sparse basis alone does not help: it trails dense on 15 of the 18 main-table comparisons and leads by at most 0.012 on the other three (SAE-unweighted in Table~\ref{tab:ablationmain}).
The learned basis directions by SAE matters too for this coordinate system: a random dictionary of the same width and sparsity, with the same idf, stays below dense on every task (Table~\ref{tab:ablationmain}; its unweighted variant and the bootstrap tests are in Appendix~\ref{app:controls}).
Stronger post-hoc fixes of the dense state, from whitening to an idf over dense coordinates, also trail \method on most tasks (Table~\ref{tab:controls}): rarity weighting needs learned features.

\subsection{Layer, width, sparsity, model size and model family}
\label{sec:sweeps}

We sweep the other configurations in \method on Lean~4 $\to$ HOL Light and Metamath (Figure~\ref{fig:ablation}; Appendix~\ref{app:ablation}).
Across all 39 configurations, including different SAE layers, SAE width, sparsity levels and model families and sizes, \method consistently beats both dense and \pcone on both tasks, by +0.18 to +0.55 MRR over dense on HOL Light and +0.08 to +0.58 on Metamath, and beats SAE-unweighted in all 78 runs, by +0.11 to +0.66.
At nearly fixed sparsity, \method rises with every doubling of the dictionary on HOL Light (0.817 at 16k to 0.935 at 1M) and levels off from 65k on Metamath, while SAE-unweighted drifts down as it widens. The same sweeps on Common-KG and MultiFarm agree (Figure~\ref{fig:sweepkg}).
The effect holds from 2B to 27B parameters (Gemma-2-27B: 0.939 and 0.738 vs.\ dense 0.41 and 0.20) and for a second model family: with Llama-3.1-8B and Llama Scope, \method beats dense in all six miniF2F configurations (Table~\ref{tab:llama}).

\subsection{Feature-level analysis}
\label{sec:interp}

\begin{figure}[t]
  \centering
  \includegraphics[width=\columnwidth]{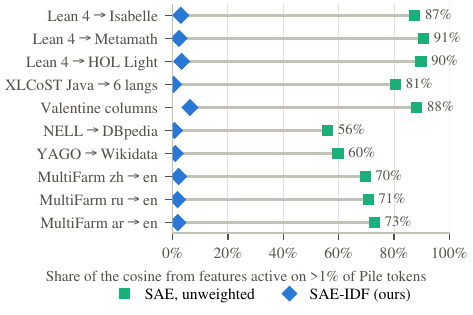}
  \caption{Share of each correct match's cosine carried by features that fire on more than 1\% of Pile tokens (Gemma-2-9B; 2,315 correct \method matches on ten tasks, and the same pairs unweighted). Without weighting (squares) these ubiquitous features carry 56--91\% of the cosine on every task; with rarity weighting (diamonds) they carry 0--6\%.}
  \label{fig:interp}
\end{figure}

\begin{table*}[t]
  \centering\footnotesize
  \setlength{\tabcolsep}{4pt}
  \begin{tabular}{@{}>{\raggedright\arraybackslash}p{0.16\textwidth}>{\raggedright\arraybackslash}p{0.31\textwidth}>{\raggedright\arraybackslash}p{0.08\textwidth}>{\raggedright\arraybackslash}p{0.17\textwidth}>{\raggedright\arraybackslash}p{0.17\textwidth}@{}}
    \toprule
    & & & \multicolumn{2}{c}{Top-1 feature by contribution to the cosine} \\
    \cmidrule(l){4-5}
    Task & Source $\to$ gold counterpart & Gold's rank & SAE-unweighted & \method \\
    \midrule
    MultiFarm zh $\to$ en & \zh{费用}{\emph{feiyong}} (fee) $\to$ ``fee for conference'' & 13 / 40 / 1 & ``mathematical or programming notation'' (53\%) & ``fees and payment structures'' (88\%) \\
    YAGO $\to$ Wikidata & Mass (no instances) $\to$ mass (instances: Earth mass, quark mass) & 7 / 24 / 1 & ``terms related to mass'' (50\%) & ``terms related to mass'' (91\%) \\
    Lean~4 $\to$ HOL Light & \texttt{Int.floor (10 * n) + \ldots{} + Int.floor (10000 * n) = 3702}\newline $\to$ \texttt{floor (\&10 * n) + \ldots{} + floor (\&10000 * n) = \&3702} & 7 / 131 / 1 & ``mathematical or programming notation'' (33\%) & ``floor and rounding operations'' (9\%) \\
    Lean~4 $\to$ Metamath & \texttt{1 / Real.cos x + Real.tan x = 22 / 7} \ldots\newline $\to$ \texttt{( ( 1 / ( cos \textasciigrave{} X ) ) + ( tan \textasciigrave{} X ) ) = ( ; 2 2 / 7 )} \ldots & 5 / 41 / 1 & ``mathematical or programming notation'' (29\%) & ``limits and trigonometric functions'' (18\%) \\
    \midrule
    \emph{Failure}: MultiFarm zh $\to$ en & \zh{有名}{\emph{you ming}} (``has first name'') $\to$ ``has first name''; ranks ``has name'' first & 1 / 2 / 2 & ``open-source license text'' (46\%) & ``variable names in programming code'' (41\%) \\
    \emph{Failure}: Valentine (ChEMBL) & \texttt{src\_id} (its only value is 1) $\to$ \texttt{SRI}; ranks \texttt{CONFSC} first & 1 / 1 / 3 & ``numbers, dates and measurements'' (84\%) & ``positive integers'' (53\%) \\
    \bottomrule
  \end{tabular}
  \caption{Case studies selected by rule, not by hand: per task, correct \method matches that Dense ranks 5th or worse, the shortest texts with the widest \method margin, and cases where Dense is right and \method is not. Gold's rank: under Dense / SAE-unweighted / \method. The last two columns give each code's top feature by contribution to the source--gold cosine, with its share; feature names are abbreviated Neuronpedia explanations. In both failures the wrong first-ranked candidate has the same top feature under \method. Selection rules: Appendix~\ref{app:cases}.}
  \label{tab:cases}
\end{table*}

We decomposed up to 300 correct \method matches per task on the ten tasks other than Bio-ML and Anatomy (2,315 in all), and the same pairs unweighted.
Each feature is described from outside our pipeline: Neuronpedia \citep{lin2023neuronpedia} publishes an automatic explanation \citep{bills2023language} for 129,642 of the 131,072 features we used.
An LLM sorted the 5,252 explanations that occur into \emph{concept}, \emph{notation}, \emph{surface} (a particular word or script) and \emph{generic} from their text alone; a second, unrelated model agreed on 92\% ($\kappa=0.80$).

Without weighting, every task matches on the same few features.
One feature, ``mathematical or programming notation'' (f106376, active on 99\% of Pile tokens; \citealp{gao2020pile}), is among SAE-unweighted's five largest contributions in 99\% of all correct matches (96--100\% per task) and in none under \method; two further ubiquitous features behave the same way (Appendix~\ref{app:cases}).
Features active on more than 1\% of the Pile carry 56--91\% of SAE-unweighted's cosine and 0--6\% of \method's (Figure~\ref{fig:interp}; per-task details in Table~\ref{tab:interp}, Appendix~\ref{app:cases}).
With weighting, two regimes appear.
On knowledge-graph and ontology terms one or two rare concept features usually carry a match (87--94\% of the cosine from features active on $<$0.1\% of the Pile); on MultiFarm a quarter of it is \emph{surface} features firing on both the foreign term and its English counterpart, a translation inside the model's features \citep{templeton2024scaling, brinkmann2025grammatical}.
Formal statements, programs and table columns spread half the cosine over 10--21 features, about half of it \emph{notation} on formal mathematics.

\paragraph{Case studies.}
In Table~\ref{tab:cases}, the Chinese MultiFarm term for \emph{fee} shares no word with ``fee for conference'', and dense ranks this gold 13th.
Without weighting, the largest contribution to their cosine comes from ``mathematical or programming notation'' (53\%), which every pair shares, and the gold falls to rank 40; with rarity weighting, ``fees and payment structures'' contributes 88\% and the gold ranks first.
The HOL Light floor problem shows the same shift: notation leads without weighting (gold at rank 131), ``floor and rounding operations'' with it (rank 1).
Failures are as readable: ``has first name'' loses to ``has name'' because both have the same top feature, ``variable names in programming code'', and the column \texttt{src\_id}, whose only value is 1, meets the gold \texttt{SRI} only at ``positive integers'', which the numeric column \texttt{CONFSC} shares as well.

\begin{table*}[t]
  \centering\small
  \begin{tabular}{@{}lrrrrrr@{}}
\toprule
Method & \makecell{Mutual-top-1\\precision} & Coverage & \makecell{Transitivity\\(triples)} & \makecell{Merge\\purity} & ARI & \makecell{Components\\(489 true)} \\
\midrule
Dense & 0.900 & 0.340 & 0.963 (872) & 0.979 & 0.519 & 1,096 \\
Dense$_{-\mathrm{PC1}}$ & 0.910 & 0.483 & 0.960 (1,868) & 0.958 & 0.646 & 895 \\
SAE unweighted & 0.616 & 0.299 & 0.902 (572) & 0.958 & 0.432 & 1,143 \\
\textbf{SAE-IDF} & \textbf{0.983} & \textbf{0.639} & \textbf{0.981} (3,474) & \textbf{0.981} & \textbf{0.795} & \textbf{758} \\
\bottomrule
\end{tabular}

  \caption{Merging the four miniF2F provers (Gemma-2-9B, layer 20; 489 problems) by linking mutual top-1 matches over the twelve ordered prover pairs. Precision and coverage: means over the pairs; in parentheses, chained triples behind each transitivity; ARI: adjusted Rand index of the merged components against the true problems. Bold: best per column (components: closest to 489).}
  \label{tab:provermerge}
\end{table*}

\section{Discussion}
\label{sec:discussion}

\paragraph{Optimal choice between neural and symbolic representations.}
A neuro-symbolic system must decide which knowledge to leave to its symbolic component \citep{yang2026generative}. Rules are exact where identity is fixed by definition (equal keys, units, formats), while neural representations suit graded similarity, the ``family resemblance'' of \citet{wittgenstein1953philosophical}.
Splitting the 551 Valentine pairs by the kind of value a column holds (Appendix~\ref{app:valentine}), \method leads dense in every class, with top-1 accuracy of 0.79--0.81 on free-text, categorical and identifier columns but 0.67 on numeric ones, the hardest class for every method.
This indicates that bare numbers, which give the features the least to match, are where rules on ranges, units and value distributions should take over, a hypothesis for future work.

\paragraph{Composition and abstention in symbolic system merging.}
Merging many systems needs pairwise alignments that agree (A to B then B to C equals A to C) and links that group symbols by their true identity.
Over the twelve ordered pairs among the four provers, with mutual top-1 matches as links, \method links almost twice as many statements as dense (coverage 0.64 vs.\ 0.34) at equal or higher consistency (transitivity 0.981 vs.\ 0.963; purity 0.981 vs.\ 0.979), and its merge comes closest to the true grouping (adjusted Rand index 0.795; dense 0.519, \pcone 0.646; Table~\ref{tab:provermerge}, Appendix~\ref{app:ablation}).
A shared space alone does not ensure this: the unweighted code, in the same space, is the least consistent (transitivity 0.902, purity 0.958).
A merge also needs abstention, since under transitive closure one false link fuses two correct clusters.
For answering selectively, the cosine margin between the first two candidates is a usable confidence: at 90\% precision \method answers 24\% of Metamath and 93\% of HOL Light queries, against 1\% and 24\% for dense (Appendix~\ref{app:abstention}).

\section{Related Work}
\label{sec:related}

\paragraph{Alignment within one kind of system.}
Ontology matching has a yearly evaluation (OAEI) and many systems, from logic-based lexical matchers such as LogMap \citep{jimenezruiz2011logmap} to fine-tuned language models \citep{he2022bertmap}, prompted LLMs \citep{hertling2023olala} and LLM agents \citep{qiang2025agentom}; \citet{euzenat2013ontology} survey it.
Schema matching \citep{rahm2001survey, koutras2021valentine}, entity alignment from seed pairs \citep{sun2020benchmarking} and cross-language code retrieval \citep{feng2020codebert, guo2022unixcoder} each have their own methods and benchmarks; none runs unchanged on another kind of system (Section~\ref{sec:baselines}).

\paragraph{Sparse autoencoders.}
SAEs recover sparse, interpretable features from superposed activations \citep{elhage2022toy, bricken2023monosemanticity, cunningham2024sparse, templeton2024scaling, gao2025scaling}, are released as open suites \citep{lieberum2024gemmascope, he2024llamascope}, and give interpretable axes for semantic search when trained on text embeddings \citep{oneill2024disentangling}.
Against doubts about their downstream use \citep{kantamneni2025sparse}, they help here once the code is rarity-weighted.

\paragraph{Neuro-symbolic models.}
Neuro-symbolic systems combine learned representations with explicit symbolic structure \citep{kautz2022third}: differentiable logic and proving (Logic Tensor Networks, \citealp{badreddine2022ltn}; neural theorem provers, \citealp{rocktaschel2017ntp}; NLProlog, \citealp{weber2019nlprolog}; $\partial$ILP, \citealp{evans2018dilp}), probabilistic logic programs over neural predicates (DeepProbLog, \citealp{manhaeve2018deepproblog}), symbolic programs run over neural scene parses (NS-CL, \citealp{mao2019nscl}), and language models paired with symbolic solvers or deduction engines (Logic-LM, \citealp{pan2023logiclm}; AlphaGeometry, \citealp{trinh2024alphageometry}).
Each reasons inside formalisms chosen for its task; none aligns the symbols of independently built systems, the step a universal symbolic backbone needs \citep{yang2026generative} and the one we address.

\section{Conclusion}
\label{sec:conclusion}

\method matches symbols across heterogeneous systems without training: native context, an LLM's sparse autoencoder, rarity weighting and cosine.
On 12 tasks across six kinds of system it beats the same model's dense state on every task with Gemma-2-9B and all but one with Llama-3.1-8B; both the dictionary and the weighting are needed, and each match reads as a few rare, named concepts.
An LLM's feature dictionary is thus an interpretable common ground for aligning symbolic systems, a step toward a universal backbone.

\section*{Limitations}

\emph{What the model can read.} The method is only as general as the base model's reading of each notation and the native context each system offers: it is at chance on SMILES, IUPAC names and InChI, on par with word TF-IDF when two notations share their identifiers (XLCoST), short of context on MultiFarm, Anatomy and label-only knowledge-graph classes, and we dropped context-less symbols such as ICD codes and drug names.
\emph{Protocol.} Every task ranks a given candidate pool in which the query has a counterpart, with one layer per model. A real merging pipeline must also decide how many counterparts a symbol has---none, one or several---under the structural constraints of the merged systems, and be tested by merging many systems with transitivity as the criterion; our margin-based abstention covers only queries that have a counterpart, and such full alignments are left to future work.
\emph{Comparisons.} We compare representations, not dedicated pipelines, and claim no state of the art on any benchmark. Trained retrievers lead on formal mathematics, where memorized aligned statements are possible; the lead over \pcone is within noise on MultiFarm and NELL $\to$ DBpedia and smaller with Llama-3.1-8B; and the idf is computed on the pool being matched (Appendix~\ref{app:controls} tests a held-out library).
\emph{Explanations.} Feature names are generated automatically and their types labelled by an LLM; they describe matches but never compute them.

\section*{Ethical Considerations}

\emph{Artifacts and licenses.} We use released models, SAEs and benchmarks for research inference and evaluation only, with no fine-tuning and no redistribution of weights or data, which is consistent with their terms: Gemma~2 under the Gemma Terms of Use and its Prohibited Use Policy, which also cover Gemma Scope (released under CC BY 4.0); Llama-3.1-8B under the Llama 3.1 Community License and its Acceptable Use Policy; Llama Scope and Qwen3-Embedding-8B under Apache-2.0; BGE-M3 under MIT.
miniF2F is released under Apache-2.0 (Lean, Isabelle), MIT (Metamath) and FreeBSD (HOL Light) licenses, miniF2F-lean4 under MIT, Bio-ML and the two Anatomy ontologies (MA, NCI Thesaurus) under CC BY 4.0; MultiFarm may be used for any scientific purpose and NCBI places no restrictions on PubChem data; XLCoST, Valentine, the OAEI Common-KG track and the Neuronpedia feature explanations are distributed by their authors for research and evaluation without a stated data license. Every artifact is cited.
\emph{Data.} The tasks consist of formal statements, programs, table schemas with sampled cell values from the public benchmark tables, and ontology and knowledge-graph terms; we collected no data from people and add no personal information.
\emph{Risks.} Automatic matching can propagate errors: under transitive closure one false link fuses two correct clusters (Section~\ref{sec:discussion}), so merges built on \method should use abstention or human review, especially for high-stakes resources such as biomedical ontologies. Feature names are generated automatically and can mislead a reader about why a match was made.

\section*{Acknowledgments}
We thank the INVITE Institute at the University of Illinois Urbana-Champaign and the ACCESS program for the computing credits that supported this work.
This material is based upon work supported by the National Science Foundation and the Institute of Education Sciences under Grant \#2229612. Any opinions, findings, and conclusions or recommendations expressed in this material are those of the author(s) and do not necessarily reflect the views of National Science Foundation or the U.S. Department of Education.
This work used the Delta system at the National Center for Supercomputing Applications [award OAC 2005572] through allocations CIS251094 and CIS261897 from the Advanced Cyberinfrastructure Coordination Ecosystem: Services \& Support (ACCESS) program \citep{boerner2023access}, which is supported by National Science Foundation grants \#2138259, \#2138286, \#2138307, \#2137603, and \#2138296.

\bibliography{references}

\begin{thebibliography}{72}
\providecommand{\natexlab}[1]{#1}

\bibitem[{Altschul et~al.(1990)Altschul, Gish, Miller, Myers, and Lipman}]{altschul1990blast}
Stephen~F. Altschul, Warren Gish, Webb Miller, Eugene~W. Myers, and David~J. Lipman. 1990.
\newblock \href {https://doi.org/10.1016/S0022-2836(05)80360-2} {Basic local alignment search tool}.
\newblock \emph{Journal of Molecular Biology}, 215(3):403--410.

\bibitem[{Arora et~al.(2017)Arora, Liang, and Ma}]{arora2017simple}
Sanjeev Arora, Yingyu Liang, and Tengyu Ma. 2017.
\newblock A simple but tough-to-beat baseline for sentence embeddings.
\newblock In \emph{International Conference on Learning Representations (ICLR)}.

\bibitem[{Ashburner et~al.(2000)Ashburner, Ball, Blake, Botstein et~al.}]{ashburner2000gene}
Michael Ashburner, Catherine~A. Ball, Judith~A. Blake, David Botstein, et~al. 2000.
\newblock \href {https://doi.org/10.1038/75556} {Gene {Ontology}: Tool for the unification of biology}.
\newblock \emph{Nature Genetics}, 25(1):25--29.

\bibitem[{Badreddine et~al.(2022)Badreddine, d'Avila Garcez, Serafini, and Spranger}]{badreddine2022ltn}
Samy Badreddine, Artur d'Avila Garcez, Luciano Serafini, and Michael Spranger. 2022.
\newblock \href {https://doi.org/10.1016/j.artint.2021.103649} {{Logic Tensor Networks}}.
\newblock \emph{Artificial Intelligence}, 303:103649.

\bibitem[{Bills et~al.(2023)Bills, Cammarata, Mossing, Tillman, Gao, Goh, Sutskever, Leike, Wu, and Saunders}]{bills2023language}
Steven Bills, Nick Cammarata, Dan Mossing, Henk Tillman, Leo Gao, Gabriel Goh, Ilya Sutskever, Jan Leike, Jeff Wu, and William Saunders. 2023.
\newblock \href {https://openaipublic.blob.core.windows.net/neuron-explainer/paper/index.html} {Language models can explain neurons in language models}.
\newblock OpenAI.

\bibitem[{Bodenreider et~al.(2005)Bodenreider, Hayamizu, Ringwald, De~Coronado, and Zhang}]{bodenreider2005mice}
Olivier Bodenreider, Terry~F. Hayamizu, Martin Ringwald, Sherri De~Coronado, and Songmao Zhang. 2005.
\newblock Of mice and men: Aligning mouse and human anatomies.
\newblock In \emph{AMIA Annual Symposium Proceedings}, pages 61--65.

\bibitem[{Boerner et~al.(2023)Boerner, Deems, Furlani, Knuth, and Towns}]{boerner2023access}
Timothy~J. Boerner, Stephen Deems, Thomas~R. Furlani, Shelley~L. Knuth, and John Towns. 2023.
\newblock \href {https://doi.org/10.1145/3569951.3597559} {{ACCESS}: Advancing innovation: {NSF}'s advanced cyberinfrastructure coordination ecosystem: Services \& support}.
\newblock In \emph{Practice and Experience in Advanced Research Computing ({PEARC} '23)}, pages 173--176. ACM.

\bibitem[{Booch et~al.(2021)Booch, Fabiano, Horesh, Kate, Lenchner, Linck, Loreggia, Murugesan, Mattei, Rossi, and Srivastava}]{booch2021thinking}
Grady Booch, Francesco Fabiano, Lior Horesh, Kiran Kate, Jon Lenchner, Nick Linck, Andrea Loreggia, Keerthiram Murugesan, Nicholas Mattei, Francesca Rossi, and Biplav Srivastava. 2021.
\newblock \href {https://doi.org/10.1609/aaai.v35i17.17765} {Thinking fast and slow in {AI}}.
\newblock In \emph{Proceedings of the AAAI Conference on Artificial Intelligence}, volume~35, pages 15042--15046.

\bibitem[{Bricken et~al.(2023)Bricken, Templeton, Batson, Chen, Jermyn, Conerly, Turner, Anil, Denison, Askell, Lasenby, Wu, Kravec, Schiefer, Maxwell, Joseph, Hatfield-Dodds, Tamkin, Nguyen, McLean, Burke, Hume, Carter, Henighan, and Olah}]{bricken2023monosemanticity}
Trenton Bricken, Adly Templeton, Joshua Batson, Brian Chen, Adam Jermyn, Tom Conerly, Nick Turner, Cem Anil, Carson Denison, Amanda Askell, Robert Lasenby, Yifan Wu, Shauna Kravec, Nicholas Schiefer, Tim Maxwell, Nicholas Joseph, Zac Hatfield-Dodds, Alex Tamkin, Karina Nguyen, and 6 others. 2023.
\newblock \href {https://transformer-circuits.pub/2023/monosemantic-features/index.html} {Towards monosemanticity: Decomposing language models with dictionary learning}.
\newblock Transformer Circuits Thread.

\bibitem[{Brinkmann et~al.(2025)Brinkmann, Wendler, Bartelt, and Mueller}]{brinkmann2025grammatical}
Jannik Brinkmann, Chris Wendler, Christian Bartelt, and Aaron Mueller. 2025.
\newblock \href {https://doi.org/10.18653/v1/2025.naacl-long.312} {Large language models share representations of latent grammatical concepts across typologically diverse languages}.
\newblock In \emph{Proceedings of the 2025 Conference of the Nations of the Americas Chapter of the Association for Computational Linguistics: Human Language Technologies (Volume 1: Long Papers)}, pages 6131--6150.

\bibitem[{Chen et~al.(2024)Chen, Xiao, Zhang, Luo, Lian, and Liu}]{chen2024bge}
Jianlyu Chen, Shitao Xiao, Peitian Zhang, Kun Luo, Defu Lian, and Zheng Liu. 2024.
\newblock \href {https://doi.org/10.18653/v1/2024.findings-acl.137} {{M3-Embedding}: Multi-linguality, multi-functionality, multi-granularity text embeddings through self-knowledge distillation}.
\newblock In \emph{Findings of the Association for Computational Linguistics: ACL 2024}, pages 2318--2335.

\bibitem[{Cunningham et~al.(2024)Cunningham, Ewart, Riggs, Huben, and Sharkey}]{cunningham2024sparse}
Hoagy Cunningham, Aidan Ewart, Logan Riggs, Robert Huben, and Lee Sharkey. 2024.
\newblock \href {https://arxiv.org/abs/2309.08600} {Sparse autoencoders find highly interpretable features in language models}.
\newblock In \emph{International Conference on Learning Representations (ICLR)}.

\bibitem[{d'Avila Garcez and Lamb(2023)}]{garcez2023neurosymbolic}
Artur d'Avila Garcez and Lu{\'i}s~C. Lamb. 2023.
\newblock \href {https://doi.org/10.1007/s10462-023-10448-w} {Neurosymbolic {AI}: The 3rd wave}.
\newblock \emph{Artificial Intelligence Review}, 56(11):12387--12406.

\bibitem[{Dawid and LeCun(2024)}]{dawid2024energy}
Anna Dawid and Yann LeCun. 2024.
\newblock \href {https://doi.org/10.1088/1742-5468/ad292b} {Introduction to latent variable energy-based models: A path toward autonomous machine intelligence}.
\newblock \emph{Journal of Statistical Mechanics: Theory and Experiment}, 2024(10):104011.

\bibitem[{Elhage et~al.(2022)Elhage, Hume, Olsson, Schiefer, Henighan, Kravec, Hatfield-Dodds, Lasenby, Drain, Chen, Grosse, McCandlish, Kaplan, Amodei, Wattenberg, and Olah}]{elhage2022toy}
Nelson Elhage, Tristan Hume, Catherine Olsson, Nicholas Schiefer, Tom Henighan, Shauna Kravec, Zac Hatfield-Dodds, Robert Lasenby, Dawn Drain, Carol Chen, Roger Grosse, Sam McCandlish, Jared Kaplan, Dario Amodei, Martin Wattenberg, and Christopher Olah. 2022.
\newblock \href {https://doi.org/10.48550/arXiv.2209.10652} {Toy models of superposition}.
\newblock Transformer Circuits Thread.
\newblock ArXiv:2209.10652.

\bibitem[{Ethayarajh(2019)}]{ethayarajh2019contextual}
Kawin Ethayarajh. 2019.
\newblock \href {https://doi.org/10.18653/v1/D19-1006} {How contextual are contextualized word representations? {C}omparing the geometry of {BERT}, {ELMo}, and {GPT-2} embeddings}.
\newblock In \emph{Proceedings of the 2019 Conference on Empirical Methods in Natural Language Processing and the 9th International Joint Conference on Natural Language Processing (EMNLP-IJCNLP)}.

\bibitem[{Euzenat and Shvaiko(2013)}]{euzenat2013ontology}
J{\'e}r{\^o}me Euzenat and Pavel Shvaiko. 2013.
\newblock \href {https://doi.org/10.1007/978-3-642-38721-0} {\emph{Ontology Matching}}, 2nd edition.
\newblock Springer, Heidelberg.

\bibitem[{Evans and Grefenstette(2018)}]{evans2018dilp}
Richard Evans and Edward Grefenstette. 2018.
\newblock \href {https://doi.org/10.1613/jair.5714} {Learning explanatory rules from noisy data}.
\newblock \emph{Journal of Artificial Intelligence Research}, 61:1--64.

\bibitem[{Fallatah et~al.(2020)Fallatah, Zhang, and Hopfgartner}]{fallatah2020gold}
Omaima Fallatah, Ziqi Zhang, and Frank Hopfgartner. 2020.
\newblock \href {https://ceur-ws.org/Vol-2788/om2020_LTpaper3.pdf} {A gold standard dataset for large knowledge graphs matching}.
\newblock In \emph{Proceedings of the 15th International Workshop on Ontology Matching (OM 2020), CEUR Workshop Proceedings 2788}.

\bibitem[{Feng et~al.(2020)Feng, Guo, Tang, Duan, Feng, Gong, Shou, Qin, Liu, Jiang, and Zhou}]{feng2020codebert}
Zhangyin Feng, Daya Guo, Duyu Tang, Nan Duan, Xiaocheng Feng, Ming Gong, Linjun Shou, Bing Qin, Ting Liu, Daxin Jiang, and Ming Zhou. 2020.
\newblock \href {https://doi.org/10.18653/v1/2020.findings-emnlp.139} {{CodeBERT}: A pre-trained model for programming and natural languages}.
\newblock In \emph{Findings of the Association for Computational Linguistics: EMNLP 2020}.

\bibitem[{Gao et~al.(2020)Gao, Biderman, Black, Golding, Hoppe, Foster, Phang, He, Thite, Nabeshima, Presser, and Leahy}]{gao2020pile}
Leo Gao, Stella Biderman, Sid Black, Laurence Golding, Travis Hoppe, Charles Foster, Jason Phang, Horace He, Anish Thite, Noa Nabeshima, Shawn Presser, and Connor Leahy. 2020.
\newblock \href {https://doi.org/10.48550/arXiv.2101.00027} {The {Pile}: An 800{GB} dataset of diverse text for language modeling}.
\newblock ArXiv:2101.00027.

\bibitem[{Gao et~al.(2025)Gao, Dupr{\'e}~la Tour, Tillman, Goh, Troll, Radford, Sutskever, Leike, and Wu}]{gao2025scaling}
Leo Gao, Tom Dupr{\'e}~la Tour, Henk Tillman, Gabriel Goh, Rajan Troll, Alec Radford, Ilya Sutskever, Jan Leike, and Jeffrey Wu. 2025.
\newblock \href {https://arxiv.org/abs/2406.04093} {Scaling and evaluating sparse autoencoders}.
\newblock In \emph{International Conference on Learning Representations (ICLR)}.

\bibitem[{{Gemma Team} et~al.(2024){Gemma Team}, Riviere, Pathak, Sessa et~al.}]{gemmateam2024gemma2}
{Gemma Team}, Morgane Riviere, Shreya Pathak, Pier~Giuseppe Sessa, et~al. 2024.
\newblock \href {https://doi.org/10.48550/arXiv.2408.00118} {Gemma 2: Improving open language models at a practical size}.
\newblock ArXiv:2408.00118.

\bibitem[{Goyal and Bengio(2022)}]{goyal2022inductive}
Anirudh Goyal and Yoshua Bengio. 2022.
\newblock \href {https://doi.org/10.1098/rspa.2021.0068} {Inductive biases for deep learning of higher-level cognition}.
\newblock \emph{Proceedings of the Royal Society A}, 478(2266):20210068.

\bibitem[{Grattafiori et~al.(2024)Grattafiori, Dubey, Jauhri et~al.}]{grattafiori2024llama3}
Aaron Grattafiori, Abhimanyu Dubey, Abhinav Jauhri, et~al. 2024.
\newblock \href {https://doi.org/10.48550/arXiv.2407.21783} {The {Llama 3} herd of models}.
\newblock ArXiv:2407.21783.

\bibitem[{Guo et~al.(2022)Guo, Lu, Duan, Wang, Zhou, and Yin}]{guo2022unixcoder}
Daya Guo, Shuai Lu, Nan Duan, Yanlin Wang, Ming Zhou, and Jian Yin. 2022.
\newblock \href {https://doi.org/10.18653/v1/2022.acl-long.499} {{UniXcoder}: Unified cross-modal pre-training for code representation}.
\newblock In \emph{Proceedings of the 60th Annual Meeting of the Association for Computational Linguistics (ACL)}.

\bibitem[{He et~al.(2022{\natexlab{a}})He, Chen, Antonyrajah, and Horrocks}]{he2022bertmap}
Yuan He, Jiaoyan Chen, Denvar Antonyrajah, and Ian Horrocks. 2022{\natexlab{a}}.
\newblock \href {https://doi.org/10.1609/aaai.v36i5.20510} {{BERTMap}: A {BERT}-based ontology alignment system}.
\newblock In \emph{Proceedings of the 36th AAAI Conference on Artificial Intelligence}.

\bibitem[{He et~al.(2022{\natexlab{b}})He, Chen, Dong, Jim{\'e}nez-Ruiz, Hadian, and Horrocks}]{he2022bioml}
Yuan He, Jiaoyan Chen, Hang Dong, Ernesto Jim{\'e}nez-Ruiz, Ali Hadian, and Ian Horrocks. 2022{\natexlab{b}}.
\newblock \href {https://doi.org/10.1007/978-3-031-19433-7_33} {Machine learning-friendly biomedical datasets for equivalence and subsumption ontology matching}.
\newblock In \emph{The Semantic Web -- ISWC 2022}.

\bibitem[{He et~al.(2024)He, Shu, Ge, Chen, Wang, Zhou, Liu, Guo, Huang, Wu, Jiang, and Qiu}]{he2024llamascope}
Zhengfu He, Wentao Shu, Xuyang Ge, Lingjie Chen, Junxuan Wang, Yunhua Zhou, Frances Liu, Qipeng Guo, Xuanjing Huang, Zuxuan Wu, Yu-Gang Jiang, and Xipeng Qiu. 2024.
\newblock \href {https://doi.org/10.48550/arXiv.2410.20526} {Llama scope: Extracting millions of features from {Llama-3.1-8B} with sparse autoencoders}.
\newblock ArXiv:2410.20526.

\bibitem[{Hertling and Paulheim(2023)}]{hertling2023olala}
Sven Hertling and Heiko Paulheim. 2023.
\newblock \href {https://doi.org/10.1145/3587259.3627571} {{OLaLa}: Ontology matching with large language models}.
\newblock In \emph{Proceedings of the 12th Knowledge Capture Conference (K-CAP)}.

\bibitem[{Jim{\'e}nez-Ruiz and Cuenca~Grau(2011)}]{jimenezruiz2011logmap}
Ernesto Jim{\'e}nez-Ruiz and Bernardo Cuenca~Grau. 2011.
\newblock \href {https://doi.org/10.1007/978-3-642-25073-6_18} {{LogMap}: Logic-based and scalable ontology matching}.
\newblock In \emph{The Semantic Web -- ISWC 2011}, LNCS 7031, pages 273--288. Springer.

\bibitem[{Kahneman(2011)}]{kahneman2011thinking}
Daniel Kahneman. 2011.
\newblock \emph{Thinking, Fast and Slow}.
\newblock Farrar, Straus and Giroux, New York.

\bibitem[{Kantamneni et~al.(2025)Kantamneni, Engels, Rajamanoharan, Tegmark, and Nanda}]{kantamneni2025sparse}
Subhash Kantamneni, Joshua Engels, Senthooran Rajamanoharan, Max Tegmark, and Neel Nanda. 2025.
\newblock \href {https://arxiv.org/abs/2502.16681} {Are sparse autoencoders useful? {A} case study in sparse probing}.
\newblock In \emph{International Conference on Machine Learning (ICML)}.

\bibitem[{Kautz(2022)}]{kautz2022third}
Henry Kautz. 2022.
\newblock \href {https://doi.org/10.1002/aaai.12036} {The third {AI} summer: {AAAI} {Robert S. Engelmore} memorial lecture}.
\newblock \emph{AI Magazine}, 43(1):105--125.

\bibitem[{Kim et~al.(2023)Kim, Chen, Cheng, Gindulyte, He, He, Li, Shoemaker, Thiessen, Yu, Zaslavsky, Zhang, and Bolton}]{kim2023pubchem}
Sunghwan Kim, Jie Chen, Tiejun Cheng, Asta Gindulyte, Jia He, Siqian He, Qingliang Li, Benjamin~A. Shoemaker, Paul~A. Thiessen, Bo~Yu, Leonid Zaslavsky, Jian Zhang, and Evan~E. Bolton. 2023.
\newblock \href {https://doi.org/10.1093/nar/gkac956} {{PubChem} 2023 update}.
\newblock \emph{Nucleic Acids Research}, 51(D1):D1373--D1380.

\bibitem[{Koutras et~al.(2021)Koutras, Siachamis, Ionescu, Psarakis, Brons, Fragkoulis, Lofi, Bonifati, and Katsifodimos}]{koutras2021valentine}
Christos Koutras, George Siachamis, Andra Ionescu, Kyriakos Psarakis, Jerry Brons, Marios Fragkoulis, Christoph Lofi, Angela Bonifati, and Asterios Katsifodimos. 2021.
\newblock \href {https://doi.org/10.1109/ICDE51399.2021.00047} {Valentine: Evaluating matching techniques for dataset discovery}.
\newblock In \emph{IEEE 37th International Conference on Data Engineering (ICDE)}.

\bibitem[{Lample and Charton(2020)}]{lample2020deep}
Guillaume Lample and Fran{\c{c}}ois Charton. 2020.
\newblock \href {https://arxiv.org/abs/1912.01412} {Deep learning for symbolic mathematics}.
\newblock In \emph{International Conference on Learning Representations (ICLR)}.

\bibitem[{Lenat(1995)}]{lenat1995cyc}
Douglas~B. Lenat. 1995.
\newblock \href {https://doi.org/10.1145/219717.219745} {{CYC}: A large-scale investment in knowledge infrastructure}.
\newblock \emph{Communications of the ACM}, 38(11):33--38.

\bibitem[{Lieberum et~al.(2024)Lieberum, Rajamanoharan, Conmy, Smith, Sonnerat, Varma, Kram{\'a}r, Dragan, Shah, and Nanda}]{lieberum2024gemmascope}
Tom Lieberum, Senthooran Rajamanoharan, Arthur Conmy, Lewis Smith, Nicolas Sonnerat, Vikrant Varma, J{\'a}nos Kram{\'a}r, Anca Dragan, Rohin Shah, and Neel Nanda. 2024.
\newblock \href {https://doi.org/10.48550/arXiv.2408.05147} {Gemma scope: Open sparse autoencoders everywhere all at once on gemma 2}.
\newblock ArXiv:2408.05147.

\bibitem[{Lin(2023)}]{lin2023neuronpedia}
Johnny Lin. 2023.
\newblock \href {https://www.neuronpedia.org} {Neuronpedia: Interactive reference and tooling for analyzing neural networks}.
\newblock Software available from neuronpedia.org.

\bibitem[{Manhaeve et~al.(2018)Manhaeve, Duman{\v{c}}i{\'c}, Kimmig, Demeester, and De~Raedt}]{manhaeve2018deepproblog}
Robin Manhaeve, Sebastijan Duman{\v{c}}i{\'c}, Angelika Kimmig, Thomas Demeester, and Luc De~Raedt. 2018.
\newblock {DeepProbLog}: Neural probabilistic logic programming.
\newblock In \emph{Advances in Neural Information Processing Systems}, volume~31.

\bibitem[{Mao et~al.(2019)Mao, Gan, Kohli, Tenenbaum, and Wu}]{mao2019nscl}
Jiayuan Mao, Chuang Gan, Pushmeet Kohli, Joshua~B. Tenenbaum, and Jiajun Wu. 2019.
\newblock The {Neuro-Symbolic Concept Learner}: Interpreting scenes, words, and sentences from natural supervision.
\newblock In \emph{International Conference on Learning Representations}.

\bibitem[{Meilicke et~al.(2012)Meilicke, Garc{\'i}a-Castro, Freitas, van Hage, Montiel-Ponsoda, Ribeiro~de Azevedo, Stuckenschmidt, {\v{S}}v{\'a}b-Zamazal, Sv{\'a}tek, Tamilin, Trojahn, and Wang}]{meilicke2012multifarm}
Christian Meilicke, Ra{\'u}l Garc{\'i}a-Castro, Fred Freitas, Willem~Robert van Hage, Elena Montiel-Ponsoda, Ryan Ribeiro~de Azevedo, Heiner Stuckenschmidt, Ond{\v{r}}ej {\v{S}}v{\'a}b-Zamazal, Vojt{\v{e}}ch Sv{\'a}tek, Andrei Tamilin, C{\'a}ssia Trojahn, and Shenghui Wang. 2012.
\newblock \href {https://doi.org/10.1016/j.websem.2012.04.001} {{MultiFarm}: A benchmark for multilingual ontology matching}.
\newblock \emph{Journal of Web Semantics}, 15:62--68.

\bibitem[{Mirzadeh et~al.(2025)Mirzadeh, Alizadeh, Shahrokhi, Tuzel, Bengio, and Farajtabar}]{mirzadeh2025gsmsymbolic}
Iman Mirzadeh, Keivan Alizadeh, Hooman Shahrokhi, Oncel Tuzel, Samy Bengio, and Mehrdad Farajtabar. 2025.
\newblock \href {https://arxiv.org/abs/2410.05229} {{GSM-Symbolic}: Understanding the limitations of mathematical reasoning in large language models}.
\newblock In \emph{International Conference on Learning Representations (ICLR)}.

\bibitem[{Mu and Viswanath(2018)}]{mu2018allbutthetop}
Jiaqi Mu and Pramod Viswanath. 2018.
\newblock \href {https://arxiv.org/abs/1702.01417} {All-but-the-top: Simple and effective postprocessing for word representations}.
\newblock In \emph{International Conference on Learning Representations (ICLR)}.

\bibitem[{O'Neill et~al.(2024)O'Neill, Ye, Iyer, and Wu}]{oneill2024disentangling}
Charles O'Neill, Christine Ye, Kartheik Iyer, and John~F. Wu. 2024.
\newblock \href {https://doi.org/10.48550/arXiv.2408.00657} {Disentangling dense embeddings with sparse autoencoders}.
\newblock ArXiv:2408.00657.

\bibitem[{{OpenAI}(2026)}]{openai2026math}
{OpenAI}. 2026.
\newblock {openai/math}.
\newblock GitHub repository, \url{https://github.com/openai/math}.
\newblock Released 6 October 2026.

\bibitem[{Pan et~al.(2023)Pan, Albalak, Wang, and Wang}]{pan2023logiclm}
Liangming Pan, Alon Albalak, Xinyi Wang, and William~Yang Wang. 2023.
\newblock \href {https://doi.org/10.18653/v1/2023.findings-emnlp.248} {{Logic-LM}: Empowering large language models with symbolic solvers for faithful logical reasoning}.
\newblock In \emph{Findings of the Association for Computational Linguistics: EMNLP 2023}, pages 3806--3824.

\bibitem[{Qiang et~al.(2024)Qiang, Wang, and Taylor}]{qiang2025agentom}
Zhangcheng Qiang, Weiqing Wang, and Kerry Taylor. 2024.
\newblock \href {https://doi.org/10.14778/3712221.3712222} {{Agent-OM}: Leveraging {LLM} agents for ontology matching}.
\newblock \emph{Proceedings of the VLDB Endowment}, 18(3):516--529.

\bibitem[{Rahm and Bernstein(2001)}]{rahm2001survey}
Erhard Rahm and Philip~A. Bernstein. 2001.
\newblock \href {https://doi.org/10.1007/s007780100057} {A survey of approaches to automatic schema matching}.
\newblock \emph{The VLDB Journal}, 10(4):334--350.

\bibitem[{Rajamanoharan et~al.(2024)Rajamanoharan, Lieberum, Sonnerat, Conmy, Varma, Kram{\'a}r, and Nanda}]{rajamanoharan2024jumping}
Senthooran Rajamanoharan, Tom Lieberum, Nicolas Sonnerat, Arthur Conmy, Vikrant Varma, J{\'a}nos Kram{\'a}r, and Neel Nanda. 2024.
\newblock \href {https://doi.org/10.48550/arXiv.2407.14435} {Jumping ahead: Improving reconstruction fidelity with {JumpReLU} sparse autoencoders}.
\newblock ArXiv:2407.14435.

\bibitem[{Robertson and Zaragoza(2009)}]{robertson2009probabilistic}
Stephen Robertson and Hugo Zaragoza. 2009.
\newblock \href {https://doi.org/10.1561/1500000019} {The probabilistic relevance framework: {BM25} and beyond}.
\newblock \emph{Foundations and Trends in Information Retrieval}, 3(4):333--389.

\bibitem[{Rockt{\"a}schel and Riedel(2017)}]{rocktaschel2017ntp}
Tim Rockt{\"a}schel and Sebastian Riedel. 2017.
\newblock End-to-end differentiable proving.
\newblock In \emph{Advances in Neural Information Processing Systems}, volume~30.

\bibitem[{Segler et~al.(2018)Segler, Preuss, and Waller}]{segler2018planning}
Marwin H.~S. Segler, Mike Preuss, and Mark~P. Waller. 2018.
\newblock \href {https://doi.org/10.1038/nature25978} {Planning chemical syntheses with deep neural networks and symbolic {AI}}.
\newblock \emph{Nature}, 555(7698):604--610.

\bibitem[{Silver et~al.(2016)Silver, Huang, Maddison, Guez et~al.}]{silver2016alphago}
David Silver, Aja Huang, Chris~J. Maddison, Arthur Guez, et~al. 2016.
\newblock \href {https://doi.org/10.1038/nature16961} {Mastering the game of {Go} with deep neural networks and tree search}.
\newblock \emph{Nature}, 529(7587):484--489.

\bibitem[{Sp{\"a}rck~Jones(1972)}]{sparckjones1972statistical}
Karen Sp{\"a}rck~Jones. 1972.
\newblock \href {https://doi.org/10.1108/eb026526} {A statistical interpretation of term specificity and its application in retrieval}.
\newblock \emph{Journal of Documentation}, 28(1):11--21.

\bibitem[{Sun et~al.(2020)Sun, Zhang, Hu, Wang, Chen, Akrami, and Li}]{sun2020benchmarking}
Zequn Sun, Qingheng Zhang, Wei Hu, Chengming Wang, Muhao Chen, Farahnaz Akrami, and Chengkai Li. 2020.
\newblock \href {https://doi.org/10.14778/3407790.3407828} {A benchmarking study of embedding-based entity alignment for knowledge graphs}.
\newblock \emph{Proceedings of the VLDB Endowment}, 13(11):2326--2340.

\bibitem[{Templeton et~al.(2024)Templeton, Conerly, Marcus, Lindsey, Bricken, Chen, Pearce, Citro, Ameisen, Jones, Cunningham, Turner, McDougall, MacDiarmid, Freeman, Sumers, Rees, Batson, Jermyn, Carter, Olah, and Henighan}]{templeton2024scaling}
Adly Templeton, Tom Conerly, Jonathan Marcus, Jack Lindsey, Trenton Bricken, Brian Chen, Adam Pearce, Craig Citro, Emmanuel Ameisen, Andy Jones, Hoagy Cunningham, Nicholas~L. Turner, Callum McDougall, Monte MacDiarmid, C.~Daniel Freeman, Theodore~R. Sumers, Edward Rees, Joshua Batson, Adam Jermyn, and 3 others. 2024.
\newblock \href {https://transformer-circuits.pub/2024/scaling-monosemanticity/index.html} {Scaling monosemanticity: Extracting interpretable features from {Claude 3 Sonnet}}.
\newblock Transformer Circuits Thread.

\bibitem[{{The mathlib Community}(2020)}]{mathlib2020}
{The mathlib Community}. 2020.
\newblock \href {https://doi.org/10.1145/3372885.3373824} {The {Lean} mathematical library}.
\newblock In \emph{Proceedings of the 9th ACM SIGPLAN International Conference on Certified Programs and Proofs (CPP)}, pages 367--381.

\bibitem[{Trinh et~al.(2024)Trinh, Wu, Le, He, and Luong}]{trinh2024alphageometry}
Trieu~H. Trinh, Yuhuai Wu, Quoc~V. Le, He~He, and Thang Luong. 2024.
\newblock \href {https://doi.org/10.1038/s41586-023-06747-5} {Solving olympiad geometry without human demonstrations}.
\newblock \emph{Nature}, 625(7995):476--482.

\bibitem[{Vaswani et~al.(2017)Vaswani, Shazeer, Parmar, Uszkoreit, Jones, Gomez, Kaiser, and Polosukhin}]{attention}
Ashish Vaswani, Noam Shazeer, Niki Parmar, Jakob Uszkoreit, Llion Jones, Aidan~N. Gomez, {\L}ukasz Kaiser, and Illia Polosukhin. 2017.
\newblock Attention is all you need.
\newblock In \emph{Advances in Neural Information Processing Systems}, volume~30.

\bibitem[{Weber et~al.(2019)Weber, Minervini, M{\"u}nchmeyer, Leser, and Rockt{\"a}schel}]{weber2019nlprolog}
Leon Weber, Pasquale Minervini, Jannes M{\"u}nchmeyer, Ulf Leser, and Tim Rockt{\"a}schel. 2019.
\newblock \href {https://doi.org/10.18653/v1/P19-1618} {{NLProlog}: Reasoning with weak unification for question answering in natural language}.
\newblock In \emph{Proceedings of the 57th Annual Meeting of the Association for Computational Linguistics (ACL)}.

\bibitem[{Wittgenstein(1953)}]{wittgenstein1953philosophical}
Ludwig Wittgenstein. 1953.
\newblock \emph{Philosophical Investigations}.
\newblock Blackwell, Oxford.

\bibitem[{Wu et~al.(2024)Wu, Qiu, Ross, Aky{\"u}rek, Chen, Wang, Kim, Andreas, and Kim}]{wu2024reasoning}
Zhaofeng Wu, Linlu Qiu, Alexis Ross, Ekin Aky{\"u}rek, Boyuan Chen, Bailin Wang, Najoung Kim, Jacob Andreas, and Yoon Kim. 2024.
\newblock \href {https://doi.org/10.18653/v1/2024.naacl-long.102} {Reasoning or reciting? exploring the capabilities and limitations of language models through counterfactual tasks}.
\newblock In \emph{Proceedings of the 2024 Conference of the North American Chapter of the Association for Computational Linguistics: Human Language Technologies (Volume 1: Long Papers)}, pages 1819--1862.

\bibitem[{Yang et~al.(2025)Yang, Li, Yang et~al.}]{yang2025qwen3}
An~Yang, Anfeng Li, Baosong Yang, et~al. 2025.
\newblock \href {https://doi.org/10.48550/arXiv.2505.09388} {Qwen3 technical report}.
\newblock \emph{Preprint}, arXiv:2505.09388.

\bibitem[{Yang(2023)}]{yang2023minif2flean4}
Kaiyu Yang. 2023.
\newblock \href {https://github.com/yangky11/miniF2F-lean4} {{miniF2F-lean4}: {Lean 4} formalization of {miniF2F}}.
\newblock GitHub repository.

\bibitem[{Yang(2026)}]{yang2026generative}
Xiaocong Yang. 2026.
\newblock \href {https://doi.org/10.48550/arXiv.2609.13529} {Generative interpretability via scalable neuro-symbolic models}.
\newblock ArXiv:2609.13529. ACM AI Summit 2026.

\bibitem[{Yang et~al.(2024)Yang, Lin, Wang, and Zhai}]{yang2024analogy}
Xiaocong Yang, Jiacheng Lin, Ziqi Wang, and Chengxiang Zhai. 2024.
\newblock \href {https://doi.org/10.48550/arXiv.2411.16454} {Learning by analogy: Enhancing few-shot prompting for math word problem solving with computational graph-based retrieval}.
\newblock \emph{Preprint}, arXiv:2411.16454.

\bibitem[{Zhang et~al.(2025)Zhang, Li, Long, Zhang, Lin, Yang, Xie, Yang, Liu, Lin, Huang, and Zhou}]{zhang2025qwen3embedding}
Yanzhao Zhang, Mingxin Li, Dingkun Long, Xin Zhang, Huan Lin, Baosong Yang, Pengjun Xie, An~Yang, Dayiheng Liu, Junyang Lin, Fei Huang, and Jingren Zhou. 2025.
\newblock \href {https://doi.org/10.48550/arXiv.2506.05176} {Qwen3 embedding: Advancing text embedding and reranking through foundation models}.
\newblock \emph{arXiv preprint arXiv:2506.05176}.

\bibitem[{Zhao et~al.(2024)Zhao, Chen, Yang, Liu, Deng, Cai, Wang, Yin, and Du}]{zhao2024explainability}
Haiyan Zhao, Hanjie Chen, Fan Yang, Ninghao Liu, Huiqi Deng, Hengyi Cai, Shuaiqiang Wang, Dawei Yin, and Mengnan Du. 2024.
\newblock \href {https://doi.org/10.1145/3639372} {Explainability for large language models: A survey}.
\newblock \emph{ACM Transactions on Intelligent Systems and Technology}, 15(2):1--38.

\bibitem[{Zheng et~al.(2022)Zheng, Han, and Polu}]{zheng2022minif2f}
Kunhao Zheng, Jesse~Michael Han, and Stanislas Polu. 2022.
\newblock \href {https://arxiv.org/abs/2109.00110} {{miniF2F}: A cross-system benchmark for formal {O}lympiad-level mathematics}.
\newblock In \emph{International Conference on Learning Representations (ICLR)}.

\bibitem[{Zhu et~al.(2022)Zhu, Jain, Suresh, Ravindran, Tipirneni, and Reddy}]{zhu2022xlcost}
Ming Zhu, Aneesh Jain, Karthik Suresh, Roshan Ravindran, Sindhu Tipirneni, and Chandan~K. Reddy. 2022.
\newblock \href {https://doi.org/10.48550/arXiv.2206.08474} {{XLCoST}: A benchmark dataset for cross-lingual code intelligence}.
\newblock ArXiv:2206.08474.

\end{thebibliography}

\appendix
\raggedbottom
\renewcommand{\dbltopfraction}{0.95}\renewcommand{\dblfloatpagefraction}{0.5}\setcounter{dbltopnumber}{3}\renewcommand{\textfraction}{0.05}

\section{Benchmark and protocol details}
\label{app:benchmarks}
\begin{table*}[tp]
  \centering\small
  \begin{tabular}{@{}p{0.13\textwidth}p{0.34\textwidth}p{0.19\textwidth}p{0.1\textwidth}p{0.12\textwidth}@{}}
    \toprule
    Kind of system & Task & Native context per symbol & Queries & Candidates per query \\
    \midrule
    Formal mathematics & miniF2F \citep{zheng2022minif2f}: Lean~4 $\to$ Isabelle / Metamath / HOL Light & the statement itself & 483 / 482 / 317 & all statements \\
    Code & XLCoST \citep{zhu2022xlcost}: Java, C++, Python, C\#, JavaScript, PHP, C $\to$ the other six & the program itself & 283--3,918 per language & official pool \\
    Table schemas & Valentine \citep{koutras2021valentine}: source $\to$ target columns & column name + 16 cell values & 551 table pairs & all target columns \\
    Knowledge graphs & OAEI Common-KG \citep{fallatah2020gold}: NELL $\to$ DBpedia, YAGO $\to$ Wikidata & up to 16 of the class's instances & 129 / 304 & 137 / 304 classes \\
    Cross-lingual ontologies & MultiFarm \citep{meilicke2012multifarm}: Chinese / Russian / Arabic $\to$ English & the term's own axioms, in its language & 246 / 246 / 239 & all target terms \\
    Biomedical ontologies & Bio-ML \citep{he2022bioml}: NCIT $\to$ DOID (validation split); OAEI Anatomy \citep{bodenreider2005mice}: mouse $\to$ human & definitions, parents and part-of relations & 634 / 1,516 & 100 per query / 3,298 classes \\
    \bottomrule
  \end{tabular}
  \caption{The main-table tasks. Appendix task: PubChem notations \citep{kim2023pubchem}.}
  \label{tab:benchmarks}
\end{table*}

\paragraph{Models and SAEs.}
Gemma-2-9B (42 layers, $d=3{,}584$), Gemma-2-2B (26 layers, $d=2{,}304$) and Gemma-2-27B (46 layers, $d=4{,}608$) are run in bfloat16; the residual stream after block $\ell$ (\texttt{hidden\_states[$\ell$+1]}) is encoded with the Gemma Scope \texttt{gemma-scope-9b-pt-res} (or \texttt{-2b-pt-res}, \texttt{-27b-pt-res}, the latter released for layers 10, 22 and 34) JumpReLU SAEs ($E(h)=\mathrm{JumpReLU}(W_{\mathrm{enc}}h+b_{\mathrm{enc}})$, Section~\ref{sec:method}, with the released per-feature thresholds $\theta_f$).
The main-table SAE is \texttt{layer\_20/width\_131k/average\_l0\_114}, the canonical release, which is the one with Neuronpedia explanations.
Llama-3.1-8B (32 layers, $d=4{,}096$) uses the Llama Scope residual SAEs \texttt{Llama3\_1-8B-Base-L\{8,12,15,20,24\}R-32x} (131k features) and \texttt{L15R-8x} (32k), applied as published, with dataset-wise input scaling and the released JumpReLU threshold.
Texts are truncated to 1,024 tokens (15 of 3,918 Java programs were truncated).

\paragraph{Views.}
Common-KG: up to 16 instances per class sampled with a fixed seed, written \texttt{Instance: <label>. Class: <label>}; a class with no instances gets the single view \texttt{Class: <label>}.
MultiFarm: one view per axiom of the term in its own ontology and language (parents and children, domain and range, properties whose domain or range it is, sub- and super-properties, inverses, disjoint classes), written \texttt{<relation> <other>. Term: <label>}.
Bio-ML and Anatomy: definitions, parents and part-of relations in the same form.
Valentine: \texttt{<table>.<column>: <v1>; <v2>; \dots} with 16 sampled values.
Only the symbol's final occurrence is pooled, and no synonyms are used anywhere.

\paragraph{Reference collections for idf.}
miniF2F and XLCoST: the candidate pool (the target prover's statements; the test file of the query language).
Valentine: the target columns of each table pair.
Common-KG, MultiFarm, Bio-ML, Anatomy: all symbols of both systems.
$\df(f)$ counts symbols, not tokens or views.

\paragraph{String baselines.}
For formal statements and PubChem the ``Strings'' column is the best of word TF-IDF, character 3--5-gram TF-IDF and BM25 over the statements: BM25 on Isabelle (0.825; word TF-IDF 0.772), word TF-IDF on Metamath (0.615; BM25 0.187) and character $n$-grams on HOL Light (0.741; word TF-IDF 0.669).
For XLCoST the official evaluator scores the word TF-IDF ranking.
For ontology and knowledge-graph terms it is the better of character 3--5-gram TF-IDF over labels and word TF-IDF over the view texts.

\paragraph{Compute.}
One 48\,GB A6000 (or two 24\,GB A5000s with the model split across them) runs Gemma-2-9B plus the 131k SAE encoder ($131{,}072\times 3{,}584$ in bfloat16, about 0.9\,GB); Llama-3.1-8B ran on one 40\,GB A100 and Gemma-2-27B on two.
Encoding takes minutes for miniF2F, Common-KG and MultiFarm and a few hours for XLCoST and Valentine; scoring and all analyses run on CPU.

\section{XLCoST per language}
\label{app:xlcost}
\begin{table*}[tp]
  \centering\small
  \begin{tabular}{@{}lrrrrrrr@{}}
\toprule
Query language & Programs & Metric & Strings & Dense & $-$PC1 & SAE unw. & \textbf{SAE-IDF} \\
\midrule
Java & 3,918 & MRR & \textbf{0.485} & 0.466 & 0.466 & 0.458 & 0.480 \\
 &  & P@6 & \textbf{0.703} & 0.582 & 0.580 & 0.564 & 0.681 \\
C++ & 3,895 & MRR & \textbf{0.486} & 0.470 & 0.469 & 0.465 & 0.478 \\
 &  & P@6 & \textbf{0.709} & 0.626 & 0.625 & 0.627 & 0.675 \\
Python & 3,853 & MRR & 0.484 & 0.474 & 0.468 & 0.470 & \textbf{0.485} \\
 &  & P@6 & \textbf{0.714} & 0.691 & 0.671 & 0.670 & 0.711 \\
C\# & 3,898 & MRR & \textbf{0.485} & 0.466 & 0.466 & 0.460 & 0.479 \\
 &  & P@6 & \textbf{0.705} & 0.591 & 0.594 & 0.599 & 0.683 \\
JavaScript & 3,851 & MRR & \textbf{0.487} & 0.475 & 0.471 & 0.469 & 0.482 \\
 &  & P@6 & \textbf{0.714} & 0.671 & 0.664 & 0.659 & 0.696 \\
PHP & 1,566 & MRR & 0.441 & 0.432 & 0.432 & 0.431 & \textbf{0.442} \\
 &  & P@6 & 0.811 & 0.774 & 0.780 & 0.765 & \textbf{0.818} \\
C & 283 & MRR & \textbf{0.408} & 0.392 & 0.395 & 0.393 & 0.405 \\
 &  & P@6 & 0.855 & 0.823 & 0.833 & 0.827 & \textbf{0.863} \\
\bottomrule
\end{tabular}

  \caption{XLCoST program-level code-to-code search, official metrics, per query language. Strings is word TF-IDF. \method beats Dense in every language on both metrics and is within 0.01 MRR of word TF-IDF.}
  \label{tab:xlcost}
\end{table*}

Table~\ref{tab:xlcost} breaks the XLCoST rows of Table~\ref{tab:main} down by query language; each program is matched against its versions in the other six languages.
Word TF-IDF has the best MRR in five of the seven languages, because the versions keep their identifiers, and \method is within 0.01 MRR of it everywhere and best on Python (0.485) and PHP (0.442).
On precision at six, \method gains most over Dense where the dense state is weakest, about +0.10 on Java (0.681 vs.\ 0.582) and +0.09 on C\# (0.683 vs.\ 0.591), and it is the best method on PHP and C, the two smallest pools.

\section{Valentine: layouts and column types}
\label{app:valentine}
\begin{table}[tbp]
  \centering\footnotesize\setlength{\tabcolsep}{3pt}
  \begin{tabular}{@{}p{0.46\columnwidth}rrr@{}}
    \toprule
    Layout (tokens pooled) & Dense & \makecell{SAE-\\unweighted} & \method \\
    \midrule
    \multicolumn{4}{@{}l}{\textit{F1}} \\
    Whole column, all tokens (main) & 0.533 & 0.458 & \textbf{0.614} \\
    Name + one value per view, all tokens (16 views) & 0.563 & 0.429 & \textbf{0.630} \\
    Value tokens only, name before them (16 views) & 0.570 & 0.555 & \textbf{0.608} \\
    Name tokens only, after one value (16 views) & \textbf{0.611} & 0.520 & 0.572 \\
    \addlinespace[2pt]
    \multicolumn{4}{@{}l}{\textit{MRR}} \\
    Whole column, all tokens (main) & 0.766 & 0.685 & \textbf{0.856} \\
    Name + one value per view, all tokens & 0.651 & 0.498 & \textbf{0.820} \\
    Value tokens only, name before them & 0.823 & 0.772 & \textbf{0.851} \\
    Name tokens only, after one value & 0.709 & 0.589 & \textbf{0.748} \\
    \bottomrule
  \end{tabular}
  \caption{Valentine, 551 pairs, four ways of reading a column. \method leads Dense on MRR in every layout and on F1 in all but the name-only one; taking each method's best layout, \method is still ahead (F1 0.630 vs.\ 0.611, MRR 0.856 vs.\ 0.823). SAE-unweighted is weakest in every layout.}
  \label{tab:layouts}
\end{table}
\begin{table*}[tp]
  \centering\small
  \begin{tabular}{@{}lrrrrr@{}}
    \toprule
    Column type & $n$ & Dense & $-$PC1 & SAE$_\text{unw}$ & SAE-IDF \\
    \midrule
    Free text & 930 & 0.727 & 0.714 & 0.684 & \textbf{0.806} \\
    Categorical & 2,313 & 0.629 & 0.645 & 0.491 & \textbf{0.790} \\
    Identifier & 1,216 & 0.702 & 0.718 & 0.630 & \textbf{0.805} \\
    Numeric & 3,887 & 0.567 & 0.490 & 0.489 & \textbf{0.670} \\
    \midrule
    Graded (text, categ.) & 3,243 & 0.657 & 0.665 & 0.546 & \textbf{0.795} \\
    Rule-governed (id, num.) & 5,103 & 0.599 & 0.544 & 0.522 & \textbf{0.702} \\
    \bottomrule
  \end{tabular}
  \caption{Valentine top-1 accuracy by the kind of value a column holds (whole-column layout, all 551 pairs). \method leads in every class; numeric columns are the hardest for every method, while \method matches identifiers as well as free text. Types are assigned by a pattern heuristic on the sampled values.}
  \label{tab:coltypes}
\end{table*}

Valentine pairs a source table with a target table whose columns must be matched.
Table~\ref{tab:layouts} varies what the model reads for each column: the whole column as one text (the main setting), 16 views each holding the name and one value, and pooling over only the value or only the name tokens.
Table~\ref{tab:coltypes} splits the columns by the kind of value they hold; Section~\ref{sec:discussion} uses this split to ask where rules on ranges, units and formats should take over.

F1 by dataset family under the whole-column layout, \method vs.\ Dense: ChEMBL 0.659 vs.\ 0.557, OpenData 0.568 vs.\ 0.467, TPC-DI 0.597 vs.\ 0.553, Wikidata 0.724 vs.\ 0.695, Magellan 1.0 for every method.

\section{Other tasks}
\label{app:tasks}
\begin{figure}[tbp]
  \centering
  \includegraphics[width=\columnwidth]{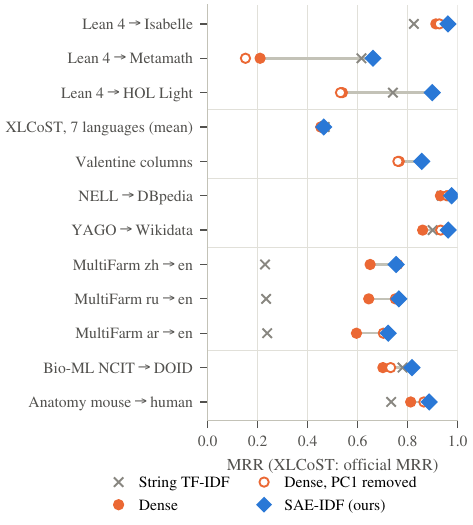}
  \caption{Main-table results with Gemma-2-9B as MRR (Valentine: MRR; XLCoST: official MRR averaged over languages). The grey bar joins Dense to \method. \method beats Dense on every task, and the gap is widest on the formal-mathematics pairs whose notation differs most from Lean~4.}
  \label{fig:main}
\end{figure}
\begin{table*}[tp]
  \centering\footnotesize
  \begin{tabular}{@{}lrrrrrr@{}}
\toprule
Task & Queries & Strings & Dense & $-$PC1 & SAE unw. & \textbf{SAE-IDF} \\
\midrule
Bio-ML NCIT $\to$ DOID, valid split & 634 & 0.779 & 0.701 & 0.732 & 0.696 & \textbf{0.817} \\
Bio-ML NCIT $\to$ DOID, train split & 3,845 & 0.789 & 0.736 & 0.758 & 0.736 & \textbf{0.815} \\
OAEI Anatomy, mouse $\to$ human & 1,516 & 0.734 & 0.812 & 0.864 & 0.801 & \textbf{0.886} \\
\midrule
PubChem SMILES $\to$ IUPAC & 1,000 & 0.009 & 0.027 & 0.012 & 0.026 & 0.015 \\
PubChem SMILES $\to$ InChI & 1,000 & 0.008 & 0.007 & 0.009 & 0.005 & 0.029 \\
PubChem IUPAC $\to$ InChI & 1,000 & 0.012 & 0.006 & 0.008 & 0.005 & 0.020 \\
\bottomrule
\end{tabular}

  \caption{Biomedical and chemical tasks with native context (MRR); Bio-ML valid and Anatomy also appear in Table~\ref{tab:main}. Bio-ML uses its 100-candidate ranking protocol; Anatomy ranks all 3,298 human classes. PubChem's chance level is 0.0075 and no method exceeds 0.03.}
  \label{tab:apptasks}
\end{table*}

Figure~\ref{fig:main} draws the Gemma-2-9B rows of Table~\ref{tab:main} as one line per task, so the size of the gain can be read directly: it is largest where the target notation differs most from Lean~4 (Metamath and HOL Light) and smallest on XLCoST, where programs share identifiers.
Table~\ref{tab:apptasks} adds the training split of Bio-ML (3,845 queries), where \method leads with 0.815 MRR against 0.789 for strings and 0.736 for Dense, and three PubChem tasks that translate one molecule between SMILES, IUPAC names and InChI.
On PubChem every method stays near chance (at most 0.029 MRR): these notations encode a structure the model does not read off the string, which is why they are not in the main table (Section~\ref{sec:datasets}).

\section{Embedding-model baseline}
\label{app:embed}
\begin{table*}[tp]
  \centering\scriptsize
  \begin{tabular}{@{}llrrrrrrrr@{}}
\toprule
 & & & \multicolumn{2}{c}{Gemma-2-9B} & \multicolumn{3}{c}{BGE-M3} & \multicolumn{2}{c}{Qwen3-Emb-8B} \\
\cmidrule(lr){4-5}\cmidrule(lr){6-8}\cmidrule(lr){9-10}
Task & Metric & Strings & Dense & \textbf{SAE-IDF} & plain & centred & $-$PC1 & plain & $-$PC1 \\
\midrule
Lean 4 $\to$ Isabelle & MRR & 0.825 & 0.912 & 0.961 & 0.956 & 0.954 & 0.932 & 0.992 & \textbf{0.995} \\
Lean 4 $\to$ Metamath & MRR & 0.615 & 0.210 & 0.661 & 0.819 & 0.591 & 0.583 & \textbf{0.972} & 0.970 \\
Lean 4 $\to$ HOL Light & MRR & 0.741 & 0.539 & 0.898 & 0.967 & 0.967 & 0.967 & 0.975 & \textbf{0.981} \\
XLCoST, 7 languages & MRR & 0.468 & 0.454 & \textbf{0.464} & 0.438 & 0.437 & 0.451 & 0.440 & 0.447 \\
XLCoST, 7 languages & P@6 & 0.745 & 0.680 & \textbf{0.732} & 0.671 & 0.669 & 0.702 & 0.684 & 0.710 \\
Valentine columns & F1 & -- & 0.533 & 0.614 & \textbf{0.645} & 0.638 & 0.612 & 0.633 & 0.609 \\
Valentine columns & MRR & -- & 0.766 & 0.856 & 0.879 & \textbf{0.890} & 0.875 & 0.872 & 0.847 \\
NELL $\to$ DBpedia & MRR & 0.936 & 0.931 & 0.976 & 0.983 & \textbf{0.988} & 0.984 & 0.973 & 0.975 \\
YAGO $\to$ Wikidata & MRR & 0.899 & 0.860 & \textbf{0.962} & 0.910 & 0.925 & 0.958 & 0.923 & 0.957 \\
MultiFarm zh $\to$ en & MRR & 0.230 & 0.650 & 0.754 & 0.716 & 0.715 & 0.743 & 0.795 & \textbf{0.809} \\
MultiFarm ru $\to$ en & MRR & 0.234 & 0.644 & 0.765 & 0.718 & 0.715 & 0.726 & 0.780 & \textbf{0.782} \\
MultiFarm ar $\to$ en & MRR & 0.238 & 0.595 & \textbf{0.722} & 0.670 & 0.681 & 0.701 & 0.690 & 0.681 \\
Bio-ML valid & MRR & 0.779 & 0.701 & \textbf{0.817} & 0.611 & 0.617 & 0.611 & 0.739 & 0.720 \\
Bio-ML train & MRR & 0.789 & 0.736 & \textbf{0.815} & 0.636 & 0.644 & 0.740 & 0.759 & 0.795 \\
Anatomy & MRR & 0.734 & 0.812 & 0.886 & 0.789 & 0.810 & 0.853 & 0.896 & \textbf{0.925} \\
\bottomrule
\end{tabular}

  \caption{Off-the-shelf retrievers against \method. Bold: best of the seven model-based columns. Plain BGE-M3 is ahead of \method on Lean~4 $\to$ Metamath, $\to$ HOL Light, Valentine and NELL $\to$ DBpedia; plain Qwen3-Embedding-8B on the formal-mathematics pairs, Valentine, MultiFarm Chinese and Russian and Anatomy. \method is ahead of every retriever column on XLCoST, YAGO $\to$ Wikidata, MultiFarm Arabic and both Bio-ML splits.}
  \label{tab:embed}
\end{table*}

BGE-M3 \citep{chen2024bge} (XLM-RoBERTa-large, CLS pooling, inputs truncated to 1,024 tokens as for the LLM) embeds exactly the view texts used elsewhere: the statement or program, the whole-column text for Valentine, and each native view for classes and terms, averaged per symbol.
Candidates and metrics are those of the main table; the centred and PC1-removed variants use the same reference collections as for the LLM hidden state (Table~\ref{tab:embed}).

Qwen3-Embedding-8B \citep{zhang2025qwen3embedding} (8B parameters, last-token pooling, bf16, the same 1,024-token truncation) was run afterwards under the same protocol, with the model's instruction format on the query side only (\texttt{Instruct: Given a symbol from one symbolic system, retrieve its counterpart in another system}); Table~\ref{tab:embed} puts it next to BGE-M3.
It is ahead of \method on 8 of the 15 rows and behind on the other 7 (narrowly on NELL $\to$ DBpedia, 0.973 vs.\ 0.976), with the pattern described in Section~\ref{sec:results}.

\section{XLCoST without the free hit}
\label{app:exclude}
\begin{table*}[tp]
  \centering\footnotesize
  \begin{tabular}{@{}lrlrrrrr@{}}
\toprule
Query language & Programs & Metric & Strings & Dense & $-$PC1 & SAE unw. & \textbf{SAE-IDF} \\
\midrule
Java & 3,918 & MRR & \textbf{0.558} & 0.528 & 0.527 & 0.516 & 0.551 \\
 &  & P@6 & \textbf{0.550} & 0.459 & 0.458 & 0.445 & 0.534 \\
C++ & 3,895 & MRR & \textbf{0.560} & 0.536 & 0.535 & 0.530 & 0.549 \\
 &  & P@6 & \textbf{0.554} & 0.496 & 0.493 & 0.494 & 0.531 \\
Python & 3,853 & MRR & 0.560 & 0.545 & 0.536 & 0.540 & \textbf{0.561} \\
 &  & P@6 & \textbf{0.557} & 0.540 & 0.528 & 0.526 & 0.553 \\
C\# & 3,898 & MRR & \textbf{0.561} & 0.530 & 0.530 & 0.522 & 0.553 \\
 &  & P@6 & \textbf{0.552} & 0.467 & 0.469 & 0.472 & 0.538 \\
JavaScript & 3,851 & MRR & \textbf{0.563} & 0.547 & 0.542 & 0.538 & 0.557 \\
 &  & P@6 & \textbf{0.558} & 0.528 & 0.522 & 0.519 & 0.545 \\
PHP & 1,566 & MRR & 0.500 & 0.488 & 0.488 & 0.487 & \textbf{0.502} \\
 &  & P@6 & 0.658 & 0.633 & 0.637 & 0.626 & \textbf{0.661} \\
C & 283 & MRR & \textbf{0.457} & 0.435 & 0.438 & 0.438 & 0.452 \\
 &  & P@6 & 0.712 & 0.699 & 0.709 & 0.695 & \textbf{0.730} \\
\bottomrule
\end{tabular}

  \caption{XLCoST program-level search with the query's own row excluded from the pool (official metric definitions otherwise). \method beats Dense in every language on both metrics; word TF-IDF stays marginally ahead on MRR in most languages.}
  \label{tab:exclude}
\end{table*}

The official XLCoST pool contains the query's own row, which every method retrieves first.
Table~\ref{tab:exclude} removes it, so the metrics count only the other-language versions of each program.
The ordering of the methods does not change: word TF-IDF is marginally ahead on MRR in five languages (by 0.005--0.011) and \method on Python and PHP, and \method beats Dense in every language on both metrics, by up to +0.075 on precision at six (Java, 0.534 vs.\ 0.459).

\section{Generated contexts}
\label{app:generated}
\begin{table*}[tp]
  \centering\footnotesize
  \begin{tabular}{@{}llrrrrr@{}}
\toprule
Task & Contexts & Strings & Dense & $-$PC1 & SAE unw. & \textbf{SAE-IDF} \\
\midrule
NELL $\to$ DBpedia & two LLMs & 0.936 & 0.964 & 0.966 & 0.916 & \textbf{0.969} \\
 & one LLM & 0.936 & 0.927 & \textbf{0.972} & 0.904 & 0.972 \\
 & native & 0.936 & 0.931 & 0.957 & 0.938 & \textbf{0.976} \\
\addlinespace[2pt]
YAGO $\to$ Wikidata & two LLMs & 0.899 & 0.908 & \textbf{0.977} & 0.915 & 0.973 \\
 & one LLM & 0.899 & 0.917 & \textbf{0.968} & 0.920 & 0.968 \\
 & native & 0.899 & 0.860 & 0.932 & 0.872 & \textbf{0.962} \\
\addlinespace[2pt]
MultiFarm zh $\to$ en & two LLMs & 0.067 & 0.780 & \textbf{0.825} & 0.492 & 0.819 \\
 & one LLM & 0.067 & 0.753 & \textbf{0.787} & 0.478 & 0.786 \\
 & native & 0.230 & 0.650 & \textbf{0.760} & 0.404 & 0.754 \\
\addlinespace[2pt]
MultiFarm ru $\to$ en & two LLMs & 0.067 & 0.703 & \textbf{0.840} & 0.270 & 0.811 \\
 & one LLM & 0.073 & 0.680 & \textbf{0.847} & 0.248 & 0.807 \\
 & native & 0.234 & 0.644 & 0.751 & 0.252 & \textbf{0.765} \\
\addlinespace[2pt]
MultiFarm ar $\to$ en & two LLMs & 0.066 & 0.588 & \textbf{0.773} & 0.195 & 0.753 \\
 & one LLM & 0.066 & 0.595 & \textbf{0.794} & 0.163 & 0.724 \\
 & native & 0.238 & 0.595 & 0.702 & 0.225 & \textbf{0.722} \\
\addlinespace[2pt]
Drug brand $\to$ generic & two LLMs & \textbf{0.762} & 0.401 & 0.464 & 0.276 & 0.448 \\
 & one LLM, shared metadata & \textbf{0.828} & 0.384 & 0.456 & 0.277 & 0.441 \\
\addlinespace[2pt]
\bottomrule
\end{tabular}

  \caption{Generated versus native contexts (MRR). ``One LLM'': both sides described by the same generator; ``two LLMs'': the source side by one generator and the target side by another, each seeing only its own system's data; ``shared metadata'': the first drug run, in which both names were given the same pharmacologic-class text. With generated text, removing the first principal component of the dense vectors matches or beats \method on every task except NELL $\to$ DBpedia with two generators (0.966 vs.\ 0.969); with native context \method leads everywhere except MultiFarm Chinese (0.754 vs.\ 0.760). Drug names have no native context and are not in the main table.}
  \label{tab:generated}
\end{table*}

Native views are the default throughout the paper. Table~\ref{tab:generated} instead describes each symbol with text that an LLM generates from its own system's data, either with one generator for both systems or with a different generator for each, so that the two sides cannot share phrasing.
Generated text raises every model-based method on MultiFarm (Dense from 0.650 to 0.780 on Chinese with two generators), and Dense$_{-\mathrm{PC1}}$ then matches or beats \method: 0.825 vs.\ 0.819 on Chinese and 0.840 vs.\ 0.811 on Russian.
A generator can also add shared text that the systems themselves do not contain, as the drug-name run with shared metadata shows (Appendix~\ref{app:tacit}); the main table therefore uses native context only, which needs no generator.

\section{Full ablation}
\label{app:ablation}
\begin{table*}[tp]
  \centering\small
  \begin{minipage}{0.92\textwidth}\begin{tabular}{@{}p{0.34\linewidth}p{0.3\linewidth}p{0.3\linewidth}@{}}
\toprule
Setting (Gemma-2-9B, layer 20, unless noted) & HOL Light & Metamath \\
\midrule
Dense baseline & 0.541 & 0.207 \\
\addlinespace[2pt]
\multicolumn{3}{@{}p{\linewidth}@{}}{\textit{SAE width} 16k / 32k / 65k / 131k / 262k / 524k / 1M, each at its canonical $L_0$ (68 / 57 / 55 / 114 / 50 / 48 / 101)} \\
\quad SAE-IDF & 0.82 / 0.86 / 0.89 / 0.90 / 0.91 / 0.93 / 0.93 & 0.62 / 0.61 / 0.71 / 0.67 / 0.68 / 0.70 / 0.66 \\
\addlinespace[2pt]
\multicolumn{3}{@{}p{\linewidth}@{}}{\textit{Sparsity}, average $L_0$ 11 / 19 / 34 / 53 / 62 / 114 / 221 / 276 (131k features)} \\
\quad SAE-IDF & 0.89 / 0.89 / 0.90 / 0.91 / 0.90 / 0.90 / 0.90 / 0.90 & 0.68 / 0.67 / 0.70 / 0.71 / 0.72 / 0.67 / 0.62 / 0.64 \\
\addlinespace[2pt]
\multicolumn{3}{@{}p{\linewidth}@{}}{\textit{Smaller model}: Gemma-2-2B, layer 12, width 16k / 65k / 262k} \\
\quad Dense & 0.462 & 0.257 \\
\quad SAE-IDF & 0.80 / 0.86 / 0.87 & 0.65 / 0.52 / 0.45 \\
\addlinespace[2pt]
\multicolumn{3}{@{}p{\linewidth}@{}}{\textit{Larger model}: Gemma-2-27B, layer 10 / 22 / 34 (131k, canonical $L_0$ 106 / 82 / 72)} \\
\quad Dense & 0.32 / 0.41 / 0.41 & 0.08 / 0.20 / 0.15 \\
\quad SAE-IDF & 0.84 / 0.94 / 0.58 & 0.41 / 0.74 / 0.60 \\
\bottomrule
\end{tabular}
\end{minipage}
  \caption{\method MRR across SAE width, sparsity and model size (Gemma Scope SAEs on Gemma-2-2B, 9B and 27B; miniF2F). Every entry beats the dense baseline in its row group. These runs were repeated separately from Table~\ref{tab:main}; at the main setting they differ from it by at most 0.006 MRR.}
  \label{tab:ablation}
\end{table*}
\begin{table*}[tp]
  \centering\small
  \begin{tabular}{@{}lrrrrrrrr@{}}
\toprule
 & \multicolumn{4}{c}{Lean 4 $\to$ HOL Light} & \multicolumn{4}{c}{Lean 4 $\to$ Metamath} \\
\cmidrule(lr){2-5}\cmidrule(lr){6-9}
Llama Scope SAE & Dense & PC1 rm. & SAE unw. & \textbf{SAE-IDF} & Dense & PC1 rm. & SAE unw. & \textbf{SAE-IDF} \\
\midrule
Layer 8, 131k & 0.502 & 0.587 & 0.436 & \textbf{0.743} & 0.500 & 0.516 & 0.567 & \textbf{0.674} \\
Layer 12, 131k & 0.509 & 0.614 & 0.451 & \textbf{0.815} & 0.630 & 0.624 & 0.582 & \textbf{0.711} \\
Layer 15, 131k & 0.603 & 0.699 & 0.611 & \textbf{0.906} & 0.646 & 0.637 & 0.675 & \textbf{0.787} \\
Layer 20, 131k & 0.304 & 0.290 & 0.209 & \textbf{0.666} & 0.269 & 0.343 & 0.369 & \textbf{0.719} \\
Layer 24, 131k & 0.291 & 0.267 & 0.274 & \textbf{0.751} & 0.124 & 0.273 & 0.202 & \textbf{0.661} \\
Layer 15, 32k & 0.603 & 0.699 & 0.583 & \textbf{0.845} & 0.646 & 0.637 & 0.626 & \textbf{0.741} \\
\bottomrule
\end{tabular}

  \caption{Second model family: Llama-3.1-8B with Llama Scope SAEs (MRR). \method beats Dense and \pcone in all six configurations on both tasks.}
  \label{tab:llama}
\end{table*}
\begin{figure*}[tp]
  \centering
  \includegraphics[width=0.95\textwidth]{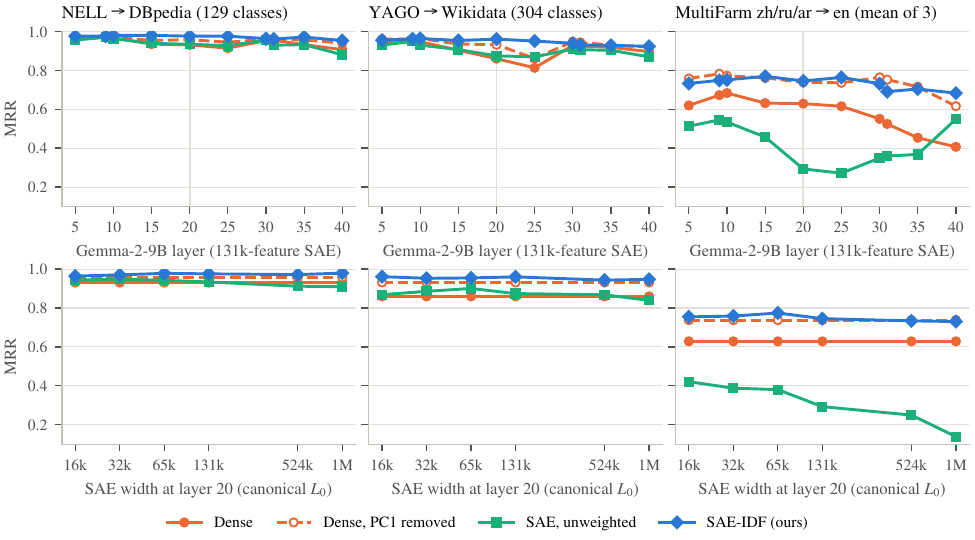}
  \caption{Layer sweep (top) and SAE-width sweep (bottom) on Common-KG and MultiFarm with native views, Gemma-2-9B. \method beats Dense and the unweighted code in every setting; \pcone is close on YAGO and MultiFarm.}
  \label{fig:sweepkg}
\end{figure*}
\begin{table*}[tp]
  \centering\footnotesize
  \begin{tabular}{@{}llrrrrr@{}}
\toprule
Task & Metric & Unweighted & \textbf{idf (collection)} & idf (Pile density) & Information & Log-odds \\
\midrule
Lean 4 $\to$ Isabelle & MRR & 0.882 & \textbf{0.961} & 0.908 & 0.740 & 0.756 \\
Lean 4 $\to$ Metamath & MRR & 0.095 & \textbf{0.661} & 0.190 & 0.102 & 0.190 \\
Lean 4 $\to$ HOL Light & MRR & 0.493 & \textbf{0.898} & 0.588 & 0.194 & 0.277 \\
XLCoST, 7 languages & MRR & 0.449 & 0.464 & 0.451 & 0.438 & \textbf{0.471} \\
Valentine columns & F1 & 0.458 & 0.614 & 0.574 & \textbf{0.615} & 0.592 \\
Valentine columns & MRR & 0.685 & \textbf{0.856} & 0.808 & 0.855 & 0.781 \\
NELL $\to$ DBpedia & MRR & 0.938 & \textbf{0.976} & 0.966 & 0.968 & 0.934 \\
YAGO $\to$ Wikidata & MRR & 0.872 & \textbf{0.962} & 0.941 & 0.947 & 0.914 \\
MultiFarm zh $\to$ en & MRR & 0.404 & \textbf{0.754} & 0.710 & 0.726 & 0.509 \\
MultiFarm ru $\to$ en & MRR & 0.252 & 0.765 & 0.699 & \textbf{0.765} & 0.319 \\
MultiFarm ar $\to$ en & MRR & 0.225 & \textbf{0.722} & 0.679 & 0.714 & 0.296 \\
Bio-ML valid & MRR & 0.696 & \textbf{0.817} & 0.793 & 0.811 & 0.760 \\
Anatomy & MRR & 0.801 & \textbf{0.886} & 0.863 & 0.885 & 0.846 \\
\bottomrule
\end{tabular}

  \caption{Alternative feature weightings on the same sparse codes (MRR unless stated). Bold: best per row.}
  \label{tab:weightings}
\end{table*}
\begin{table*}[tp]
  \centering\footnotesize
  \begin{tabular}{@{}llrrrr@{}}
\toprule
Task & Llama Scope SAE & Dense & $-$PC1 & SAE unw. & \textbf{SAE-IDF} \\
\midrule
NELL $\to$ DBpedia & layer 8, 131k & 0.953 & 0.960 & 0.945 & \textbf{0.970} \\
 & layer 12, 131k & 0.939 & 0.950 & 0.937 & \textbf{0.973} \\
 & layer 15, 131k & 0.949 & \textbf{0.960} & 0.930 & 0.958 \\
 & layer 15, 32k & 0.949 & 0.960 & 0.942 & \textbf{0.961} \\
 & layer 20, 131k & 0.947 & \textbf{0.958} & 0.904 & 0.947 \\
 & layer 24, 131k & 0.938 & \textbf{0.951} & 0.908 & 0.936 \\
YAGO $\to$ Wikidata & layer 8, 131k & 0.889 & 0.937 & 0.885 & \textbf{0.949} \\
 & layer 12, 131k & 0.806 & 0.922 & 0.861 & \textbf{0.955} \\
 & layer 15, 131k & 0.889 & 0.944 & 0.915 & \textbf{0.944} \\
 & layer 15, 32k & 0.889 & 0.944 & 0.920 & \textbf{0.955} \\
 & layer 20, 131k & 0.912 & \textbf{0.935} & 0.899 & 0.928 \\
 & layer 24, 131k & 0.915 & \textbf{0.936} & 0.875 & 0.926 \\
MultiFarm & layer 8, 131k & 0.672 & \textbf{0.751} & 0.575 & 0.739 \\
 & layer 12, 131k & 0.583 & 0.687 & 0.496 & \textbf{0.758} \\
 & layer 15, 131k & 0.633 & \textbf{0.733} & 0.556 & 0.728 \\
 & layer 15, 32k & 0.633 & \textbf{0.733} & 0.528 & 0.726 \\
 & layer 20, 131k & 0.625 & 0.716 & 0.470 & \textbf{0.722} \\
 & layer 24, 131k & 0.523 & 0.684 & 0.307 & \textbf{0.704} \\
\bottomrule
\end{tabular}

  \caption{Llama-3.1-8B layer sweep on Common-KG and MultiFarm (mean of the three pairs), native views, MRR.}
  \label{tab:llamakg}
\end{table*}
\begin{table*}[tp]
  \centering\footnotesize
  \begin{tabular}{@{}lrrrrr@{}}
\toprule
Task & Seeds & Dense & $-$PC1 & SAE unw. & \textbf{SAE-IDF} \\
\midrule
NELL $\to$ DBpedia & 4 & 0.934 $\pm$ 0.002 & 0.954 $\pm$ 0.003 & 0.933 $\pm$ 0.004 & 0.976 $\pm$ 0.000 \\
YAGO $\to$ Wikidata & 4 & 0.861 $\pm$ 0.001 & 0.932 $\pm$ 0.003 & 0.874 $\pm$ 0.002 & 0.960 $\pm$ 0.002 \\
Valentine (F1) & 3 & 0.535 $\pm$ 0.002 & 0.534 $\pm$ 0.003 & 0.459 $\pm$ 0.002 & 0.614 $\pm$ 0.001 \\
Valentine (MRR) & 3 & 0.768 $\pm$ 0.001 & 0.762 $\pm$ 0.001 & 0.685 $\pm$ 0.002 & 0.857 $\pm$ 0.002 \\
\bottomrule
\end{tabular}

  \caption{Mean $\pm$ half-range over sampling seeds (MRR unless stated). Gemma-2-9B, layer 20, 131k SAE.}
  \label{tab:seeds}
\end{table*}
\begin{table*}[tp]
  \centering\footnotesize
  \setlength{\tabcolsep}{3.8pt}
  \begin{tabular}{@{}lrrrrrrrr@{}}
\toprule
 & \multicolumn{4}{c}{Lean 4 $\to$ HOL Light} & \multicolumn{4}{c}{Lean 4 $\to$ Metamath} \\
\cmidrule(lr){2-5}\cmidrule(lr){6-9}
Configuration & Dense & PC1 rm. & SAE unw. & \textbf{SAE-IDF} & Dense & PC1 rm. & SAE unw. & \textbf{SAE-IDF} \\
\midrule
Gemma-2-9B, layer 5, 131k & 0.248 & 0.224 & 0.177 & \textbf{0.800} & 0.138 & 0.159 & 0.068 & \textbf{0.512} \\
layer 9 & 0.317 & 0.320 & 0.348 & \textbf{0.840} & 0.190 & 0.172 & 0.140 & \textbf{0.583} \\
layer 10 & 0.335 & 0.330 & 0.433 & \textbf{0.867} & 0.212 & 0.175 & 0.130 & \textbf{0.451} \\
layer 15 & 0.458 & 0.434 & 0.524 & \textbf{0.880} & 0.233 & 0.173 & 0.100 & \textbf{0.727} \\
layer 20 ($L_0$ 114, main table) & 0.541 & 0.530 & 0.493 & \textbf{0.900} & 0.207 & 0.149 & 0.092 & \textbf{0.667} \\
layer 25 & 0.439 & 0.553 & 0.280 & \textbf{0.882} & 0.179 & 0.145 & 0.112 & \textbf{0.623} \\
layer 30 & 0.214 & 0.248 & 0.144 & \textbf{0.739} & 0.123 & 0.089 & 0.088 & \textbf{0.693} \\
layer 31 & 0.210 & 0.209 & 0.122 & \textbf{0.757} & 0.121 & 0.076 & 0.077 & \textbf{0.703} \\
layer 35 & 0.225 & 0.204 & 0.115 & \textbf{0.716} & 0.100 & 0.076 & 0.078 & \textbf{0.594} \\
layer 40 & 0.173 & 0.139 & 0.080 & \textbf{0.464} & 0.077 & 0.049 & 0.039 & \textbf{0.301} \\
layer 20, 16k ($L_0$ 68) & 0.541 & 0.530 & 0.521 & \textbf{0.820} & 0.207 & 0.149 & 0.119 & \textbf{0.625} \\
layer 20, 16k ($L_0$ 58) & 0.541 & 0.531 & 0.490 & \textbf{0.817} & 0.209 & 0.151 & 0.114 & \textbf{0.621} \\
layer 20, 32k ($L_0$ 57) & 0.541 & 0.530 & 0.503 & \textbf{0.862} & 0.207 & 0.149 & 0.109 & \textbf{0.608} \\
layer 20, 65k ($L_0$ 55) & 0.541 & 0.530 & 0.417 & \textbf{0.890} & 0.207 & 0.149 & 0.086 & \textbf{0.706} \\
layer 20, 262k ($L_0$ 50) & 0.541 & 0.531 & 0.442 & \textbf{0.912} & 0.209 & 0.151 & 0.063 & \textbf{0.684} \\
layer 20, 524k ($L_0$ 48) & 0.541 & 0.530 & 0.462 & \textbf{0.927} & 0.207 & 0.149 & 0.060 & \textbf{0.698} \\
layer 20, 1M ($L_0$ 101) & 0.541 & 0.530 & 0.455 & \textbf{0.931} & 0.207 & 0.149 & 0.077 & \textbf{0.659} \\
layer 20, 1M ($L_0$ 57) & 0.541 & 0.531 & 0.436 & \textbf{0.935} & 0.209 & 0.151 & 0.059 & \textbf{0.668} \\
layer 20, 131k, $L_0$ 11 & 0.541 & 0.530 & 0.403 & \textbf{0.894} & 0.207 & 0.149 & 0.046 & \textbf{0.676} \\
$L_0$ 19 & 0.541 & 0.530 & 0.314 & \textbf{0.893} & 0.207 & 0.149 & 0.056 & \textbf{0.670} \\
$L_0$ 34 & 0.541 & 0.530 & 0.423 & \textbf{0.904} & 0.207 & 0.149 & 0.057 & \textbf{0.700} \\
$L_0$ 53 & 0.541 & 0.530 & 0.462 & \textbf{0.906} & 0.207 & 0.149 & 0.080 & \textbf{0.708} \\
$L_0$ 62 & 0.541 & 0.530 & 0.517 & \textbf{0.904} & 0.207 & 0.149 & 0.084 & \textbf{0.720} \\
$L_0$ 221 & 0.541 & 0.530 & 0.501 & \textbf{0.896} & 0.207 & 0.149 & 0.109 & \textbf{0.625} \\
$L_0$ 276 & 0.541 & 0.530 & 0.487 & \textbf{0.897} & 0.207 & 0.149 & 0.116 & \textbf{0.639} \\
Gemma-2-2B, layer 12, 16k & 0.462 & 0.487 & 0.492 & \textbf{0.800} & 0.257 & 0.238 & 0.236 & \textbf{0.651} \\
Gemma-2-2B, layer 12, 65k & 0.462 & 0.487 & 0.575 & \textbf{0.862} & 0.257 & 0.238 & 0.122 & \textbf{0.523} \\
Gemma-2-2B, layer 12, 262k & 0.462 & 0.487 & 0.491 & \textbf{0.872} & 0.257 & 0.238 & 0.156 & \textbf{0.452} \\
Llama-3.1-8B, layer 8, 131k & 0.502 & 0.587 & 0.436 & \textbf{0.743} & 0.500 & 0.516 & 0.567 & \textbf{0.674} \\
Llama-3.1-8B, layer 12, 131k & 0.509 & 0.614 & 0.451 & \textbf{0.815} & 0.630 & 0.624 & 0.582 & \textbf{0.711} \\
Llama-3.1-8B, layer 15, 131k & 0.603 & 0.699 & 0.611 & \textbf{0.906} & 0.646 & 0.637 & 0.675 & \textbf{0.787} \\
Llama-3.1-8B, layer 15, 32k & 0.603 & 0.699 & 0.583 & \textbf{0.845} & 0.646 & 0.637 & 0.626 & \textbf{0.741} \\
Llama-3.1-8B, layer 20, 131k & 0.304 & 0.290 & 0.209 & \textbf{0.666} & 0.269 & 0.343 & 0.369 & \textbf{0.719} \\
Llama-3.1-8B, layer 24, 131k & 0.291 & 0.267 & 0.274 & \textbf{0.751} & 0.124 & 0.273 & 0.202 & \textbf{0.661} \\
Gemma-2-27B, layer 10, 131k ($L_0$ 106) & 0.319 & 0.288 & 0.282 & \textbf{0.843} & 0.077 & 0.114 & 0.058 & \textbf{0.405} \\
Gemma-2-27B, layer 22, 131k ($L_0$ 82) & 0.408 & 0.328 & 0.282 & \textbf{0.939} & 0.199 & 0.097 & 0.139 & \textbf{0.738} \\
Gemma-2-27B, layer 22, 131k ($L_0$ 48) & 0.408 & 0.328 & 0.281 & \textbf{0.917} & 0.199 & 0.097 & 0.126 & \textbf{0.731} \\
Gemma-2-27B, layer 22, 131k ($L_0$ 150) & 0.408 & 0.328 & 0.280 & \textbf{0.924} & 0.199 & 0.097 & 0.172 & \textbf{0.709} \\
Gemma-2-27B, layer 34, 131k ($L_0$ 72) & 0.407 & 0.491 & 0.154 & \textbf{0.584} & 0.145 & 0.126 & 0.083 & \textbf{0.597} \\
\bottomrule
\end{tabular}

  \caption{All 39 ablation configurations (78 runs) on miniF2F Lean~4 $\to$ HOL Light and $\to$ Metamath (MRR). Gemma Scope widths are given at the sparsity used; ``$L_0$'' rows are the 131k-feature SAE at layer 20. \method beats Dense, \pcone and SAE-unweighted in every run.}
  \label{tab:ablationfull}
\end{table*}

\paragraph{Beyond formal mathematics.}
Figure~\ref{fig:sweepkg} repeats the layer and width sweeps on Common-KG and MultiFarm with the same SAE files (the 262k and matched-sparsity 16k/1M files were not available on the cluster used for these runs).
Over the 45 (task, setting) pairs \method beats dense by +0.006 to +0.28 MRR and the unweighted code in every pair; \pcone is behind in 32 of 45 (15 of 15 on NELL $\to$ DBpedia, 10 on YAGO $\to$ Wikidata, 7 on MultiFarm).
On the knowledge-graph tasks every layer from 5 to 25 is within 0.03 MRR; on MultiFarm layers 15 and 25 are best (0.769 and 0.763); the last layers fall off everywhere.
Width helps less than on HOL Light: NELL $\to$ DBpedia rises slightly (0.964 to 0.979), YAGO $\to$ Wikidata is flat, and MultiFarm peaks at 65k features (0.774) and declines to 0.730 at 1M, while the unweighted code on MultiFarm collapses as the dictionary widens (0.42 to 0.14).

\paragraph{Feature weightings.}
Every evaluator also scores the alternatives to collection idf that the pipeline computes: an idf from Neuronpedia's Pile activation densities (which needs no matched collection), an information weighting (log of in-symbol over background firing rate) and a log-odds weighting.
Collection idf is best on 10 of 13 comparisons (Table~\ref{tab:weightings}); the information weighting is close on ontology terms and tables, and the Pile-density idf is far behind on formal mathematics (0.190 vs.\ 0.661 on Metamath), where the generic features are specific to the pair of systems rather than to the Pile.

\paragraph{Four-prover merge.}
Running all twelve ordered pairs among Lean~4, Isabelle, Metamath and HOL Light (Gemma-2-9B, layer 20) and merging mutual-top-1 matches into connected components gives Table~\ref{tab:provermerge}: \method's alignments are more precise and cover almost twice as many statements as dense at equal or higher consistency, so they merge into fewer components, closer to the 489 true problems. Transitivity is computed on the chained triples a method links, so each method is measured on its own subset; \pcone removes the mean and first principal direction of all merged statements. The same procedure on XLCoST (911 programs in up to seven languages), where notations share their surface, gives transitivity 0.9987 for \method and 0.9967 for dense and an adjusted Rand index of 0.983 and 0.962 (\pcone was not rerun there).

\paragraph{Second model family on the main table.}
Table~\ref{tab:main} includes Llama-3.1-8B with the Llama Scope layer-15 131k SAE on every main-table task with the same native views.
\method beats dense on 13 of the 14 rows and \pcone on 10: the formal-mathematics gains carry over (Llama's dense state is far stronger on Metamath), Valentine improves (F1 0.625 vs.\ 0.588 dense, MRR 0.862), the knowledge-graph, cross-lingual and biomedical tasks are ties or small deficits against \pcone (NELL $\to$ DBpedia 0.958 vs.\ 0.960, MultiFarm ru $\to$ en 0.726 vs.\ 0.751, Anatomy 0.844 vs.\ 0.871), and on XLCoST \method trails dense by 0.01 MRR while leading on Precision@6.
Without the idf factor the Llama code stays below \method on 13 of the 14 rows (all but XLCoST MRR, 0.448 vs.\ 0.442), and on the Bio-ML training split \method reaches 0.789 against 0.727 for dense.

\paragraph{Llama layers on knowledge-graph and cross-lingual terms.}
Table~\ref{tab:llamakg} repeats the Common-KG and MultiFarm runs with every Llama Scope layer: the layer-15 ties with \pcone become leads at layers 8 and 12 on both knowledge-graph pairs; on MultiFarm (mean of the three language pairs) \method leads at layers 12, 20 and 24 and trails \pcone by at most 0.012 at the others. Across the 18 (task, layer) cells \method beats dense in 16 and \pcone in 10.

\paragraph{Seed study.}
The only stochastic step is the sampling of views: up to 16 instances per Common-KG class and 16 cell values per Valentine column.
Table~\ref{tab:seeds} repeats the main-table runs with other sampling seeds (four seeds for Common-KG, three for Valentine); the ranges are far smaller than the gaps between methods.

\section{Feature-level decomposition and case studies}
\label{app:cases}
\begin{table*}[tp]
  \centering\scriptsize
  \begin{tabular}{@{}lrrrrr@{}}
\toprule
Task & Matches & \makecell[r]{Common ($>$1\%):\\unw. $\to$ SAE-IDF} & \makecell[r]{Rare ($<$0.1\%)\\SAE-IDF} & \makecell[r]{Features for\\50\% / 80\%} & \makecell[r]{Concept / surface / notation\\(SAE-IDF top-5)} \\
\midrule
Lean 4 $\to$ Isabelle & 300 & 87\% $\to$ 3\% & 32\% & 21 / 146 & 43\% / 7\% / 48\% \\
Lean 4 $\to$ Metamath & 259 & 91\% $\to$ 2\% & 31\% & 15 / 88 & 39\% / 4\% / 55\% \\
Lean 4 $\to$ HOL Light & 271 & 90\% $\to$ 3\% & 30\% & 17 / 137 & 46\% / 8\% / 44\% \\
XLCoST, Java $\to$ 6 languages & 300 & 81\% $\to$ 0\% & 75\% & 10 / 88 & 51\% / 11\% / 35\% \\
Valentine columns & 300 & 88\% $\to$ 6\% & 49\% & 10 / 54 & 79\% / 4\% / 13\% \\
NELL $\to$ DBpedia & 124 & 56\% $\to$ 1\% & 94\% & 1 / 2 & 94\% / 5\% / 1\% \\
YAGO $\to$ Wikidata & 286 & 60\% $\to$ 1\% & 93\% & 1 / 3 & 87\% / 11\% / 2\% \\
MultiFarm zh $\to$ en & 162 & 70\% $\to$ 2\% & 87\% & 1 / 4 & 69\% / 26\% / 1\% \\
MultiFarm ru $\to$ en & 163 & 71\% $\to$ 2\% & 88\% & 1 / 4 & 72\% / 24\% / 2\% \\
MultiFarm ar $\to$ en & 150 & 73\% $\to$ 2\% & 87\% & 2 / 4 & 71\% / 24\% / 1\% \\
\bottomrule
\end{tabular}

  \caption{Feature-level decomposition of correct matches. \emph{Common}: share of the cosine from features active on $>$1\% of Pile tokens, under the unweighted code and under \method. \emph{Rare}: share from features active on $<$0.1\%. \emph{Features for 50\%/80\%}: median number of shared features needed to reach that share of the cosine. The last column splits \method's five largest contributions per match by explanation type.}
  \label{tab:interp}
\end{table*}

Cases were selected per task by three rules, applied to the evaluation's own rankings: \emph{wins}, correct \method matches whose gold counterpart Dense ranks 5th or worse, taking the shortest query texts with the widest \method margin over the runner-up; \emph{failures}, pairs where Dense ranks the gold counterpart first and \method does not, taking the shortest; and the features with the largest \method contributions over all decomposed matches of the task.
For each case we list the gold's rank under every method, the top five shared features with their share of the cosine and Pile density, the same pair's top shared features without weighting, and, for failures, the features shared with the wrong top candidate.
One XLCoST ``failure'' is spurious, because the dataset contains the program twice.
The full listings for the ten decomposed tasks (38 cases: 28 wins and 10 failures) are released with the code.

\section{Abstention}
\label{app:abstention}
\begin{table*}[tp]
  \centering\footnotesize
  \begin{tabular}{@{}lrrrrrrr@{}}
\toprule
 & & \multicolumn{3}{c}{Dense} & \multicolumn{3}{c}{\textbf{SAE-IDF}} \\
\cmidrule(lr){3-5}\cmidrule(lr){6-8}
Task & $n$ & Top-1 & Cov.@90\% & Cov.@95\% & Top-1 & Cov.@90\% & Cov.@95\% \\
\midrule
Lean 4 $\to$ Isabelle & 483 & 0.880 & 0.96 & 0.89 & 0.936 & 1.00 & 0.98 \\
Lean 4 $\to$ Metamath & 482 & 0.127 & 0.01 & 0.01 & 0.537 & 0.24 & 0.21 \\
Lean 4 $\to$ HOL Light & 317 & 0.454 & 0.24 & 0.20 & 0.855 & 0.93 & 0.85 \\
Valentine columns & 8,346 & 0.622 & 0.38 & 0.22 & 0.738 & 0.67 & 0.53 \\
NELL $\to$ DBpedia & 129 & 0.899 & 0.99 & 0.91 & 0.961 & 1.00 & 1.00 \\
YAGO $\to$ Wikidata & 304 & 0.819 & 0.89 & 0.76 & 0.941 & 1.00 & 0.97 \\
MultiFarm zh $\to$ en & 246 & 0.549 & 0.42 & 0.32 & 0.659 & 0.41 & 0.30 \\
MultiFarm ru $\to$ en & 246 & 0.553 & 0.29 & 0.20 & 0.663 & 0.34 & 0.20 \\
MultiFarm ar $\to$ en & 239 & 0.523 & 0.31 & 0.21 & 0.628 & 0.44 & 0.24 \\
\bottomrule
\end{tabular}

  \caption{Margin-based abstention. Cov.@90\%: the fraction of queries that can be answered while top-1 precision stays at or above 90\%.}
  \label{tab:abstention}
\end{table*}

If a matcher may abstain, how much coverage does it keep at a given precision?
For every query the confidence is the cosine margin between the best and the second-best candidate; queries are answered in decreasing order of confidence.
Table~\ref{tab:abstention} gives top-1 accuracy with every query answered and the largest coverage at which precision is still at least 90\% or 95\%, for \method and the dense hidden state, from the same weights the evaluation used (XLCoST is omitted because its official pool contains the query's own row).
Every query in these benchmarks has a counterpart, so this measures selective answering; detecting symbols that have no counterpart is not tested.

\section{Post-hoc controls and analyses}
\label{app:controls}
\begin{table*}[tp]
  \centering\scriptsize
  \setlength{\tabcolsep}{4pt}
  \begin{tabular}{@{}llrrrrrrrr@{}}
\toprule
Task & Metric & Dense & $-$PC1 & $-$36 PCs & z-score & ZCA & dense-idf & sparse dense-idf & \textbf{SAE-IDF} \\
\midrule
Lean 4 $\to$ Isabelle & MRR & 0.912 & 0.926 & 0.928 & 0.942 & 0.946 & 0.942 & 0.914 & \textbf{0.961} \\
Lean 4 $\to$ Metamath & MRR & 0.210 & 0.151 & 0.401 & 0.257 & 0.465 & 0.251 & 0.235 & \textbf{0.661} \\
Lean 4 $\to$ HOL Light & MRR & 0.539 & 0.531 & 0.728 & 0.707 & 0.786 & 0.706 & 0.605 & \textbf{0.898} \\
XLCoST, 7 languages & MRR & 0.454 & 0.452 & 0.461 & 0.456 & 0.445 & 0.456 & 0.452 & \textbf{0.464} \\
XLCoST, 7 languages & P@6 & 0.680 & 0.678 & 0.709 & 0.709 & 0.599 & 0.709 & 0.705 & \textbf{0.732} \\
Valentine columns & F1 & 0.533 & 0.534 & 0.468 & 0.460 & 0.410 & 0.399 & 0.208 & \textbf{0.614} \\
Valentine columns & MRR & 0.766 & 0.761 & 0.548 & 0.529 & 0.503 & 0.485 & 0.391 & \textbf{0.856} \\
NELL $\to$ DBpedia & MRR & 0.931 & 0.957 & 0.973 & 0.923 & \textbf{0.989} & 0.923 & 0.925 & 0.976 \\
YAGO $\to$ Wikidata & MRR & 0.860 & 0.932 & 0.957 & 0.814 & \textbf{0.964} & 0.812 & 0.847 & 0.962 \\
MultiFarm zh $\to$ en & MRR & 0.650 & \textbf{0.760} & 0.250 & 0.714 & 0.136 & 0.710 & 0.582 & 0.754 \\
MultiFarm ru $\to$ en & MRR & 0.644 & 0.751 & 0.259 & 0.713 & 0.156 & 0.710 & 0.582 & \textbf{0.765} \\
MultiFarm ar $\to$ en & MRR & 0.595 & 0.702 & 0.240 & 0.651 & 0.170 & 0.644 & 0.545 & \textbf{0.722} \\
\bottomrule
\end{tabular}

  \caption{Stronger dense controls (MRR unless stated). Bold: best per row. Whitening recovers the knowledge-graph rows and nothing else; no control comes within 0.1 MRR of \method on Metamath or HOL Light.}
  \label{tab:controls}
\end{table*}
\begin{table*}[tp]
  \centering\scriptsize
  \begin{tabular}{@{}llrrrrr@{}}
\toprule
Task & Metric & Dense & Random, unw. & Random + idf & SAE unw. & \textbf{SAE-IDF} \\
\midrule
Lean 4 $\to$ Isabelle & MRR & 0.912 & 0.845 & 0.865 & 0.882 & \textbf{0.961} \\
Lean 4 $\to$ Metamath & MRR & 0.210 & 0.156 & 0.146 & 0.095 & \textbf{0.661} \\
Lean 4 $\to$ HOL Light & MRR & 0.539 & 0.484 & 0.349 & 0.493 & \textbf{0.898} \\
XLCoST, 7 languages & MRR & 0.454 & 0.429 & 0.417 & 0.449 & \textbf{0.464} \\
XLCoST, 7 languages & P@6 & 0.680 & 0.607 & 0.627 & 0.673 & \textbf{0.732} \\
Valentine columns & F1 & 0.533 & 0.460 & 0.482 & 0.458 & \textbf{0.614} \\
Valentine columns & MRR & 0.766 & 0.654 & 0.647 & 0.685 & \textbf{0.856} \\
NELL $\to$ DBpedia & MRR & 0.931 & 0.901 & 0.901 & 0.938 & \textbf{0.976} \\
YAGO $\to$ Wikidata & MRR & 0.860 & 0.797 & 0.797 & 0.872 & \textbf{0.962} \\
MultiFarm zh $\to$ en & MRR & 0.650 & 0.482 & 0.533 & 0.404 & \textbf{0.754} \\
MultiFarm ru $\to$ en & MRR & 0.644 & 0.448 & 0.542 & 0.252 & \textbf{0.765} \\
MultiFarm ar $\to$ en & MRR & 0.595 & 0.368 & 0.425 & 0.225 & \textbf{0.722} \\
\midrule
Bio-ML valid & MRR & 0.701 & 0.680 & 0.692 & 0.696 & \textbf{0.817} \\
Bio-ML train & MRR & 0.736 & 0.693 & 0.711 & 0.736 & \textbf{0.815} \\
Anatomy & MRR & 0.812 & 0.791 & 0.788 & 0.801 & \textbf{0.886} \\
\bottomrule
\end{tabular}

  \caption{A random 131k-feature dictionary with the SAE's sparsity, unweighted and with the same idf weighting, against the SAE (MRR unless stated). Bold: best per row.}
  \label{tab:random}
\end{table*}

The experiments in this section were added after the main results, with their decision rules written down before they ran.

\paragraph{Stronger dense controls.}
On the stored dense vectors of every main-table run, each fitted on the same reference collection $D$ as the idf: removal of the top 36 principal directions ($\approx d/100$), per-dimension standardisation, ZCA whitening with Ledoit--Wolf shrinkage, and two \emph{dense-idf} variants that mark a symbol's 114 largest standardised coordinates as active, compute an idf over those activity sets, and weight either all coordinates or only the active ones (Table~\ref{tab:controls}).
The pre-registered rule---drop the anisotropy claim if any single control is within noise of \method on at least 8 of 16 rows---was not met: whitening is the strongest control and ties or beats \method only on the two Common-KG pairs, where it reaches 0.989 and 0.964, while it collapses on MultiFarm (0.14--0.17 MRR), falls below dense on XLCoST and leaves Metamath at 0.465 and HOL Light at 0.786.
For Valentine the transforms are fitted on all target columns of all pairs.

\paragraph{A random dictionary of the same width and sparsity.}
To separate the learned features from sparsity plus rarity weighting, we replaced the SAE encoder by a random one of the same width ($F=131{,}072$; i.i.d.\ Gaussian columns, unit norm, no bias) with one global JumpReLU threshold calibrated so that its mean number of active features per token on a fixed sample of symbol tokens equals the real SAE's on the same tokens (122.6 vs.\ 127.8), and ran the whole pipeline---native views, per-token encoding, averaging, idf over the same collection, cosine---unchanged on every main-table row (Table~\ref{tab:random}).
The pre-registered rule: the title stands if \method is significantly ahead of the random dictionary with idf on at least 12 of 16 rows; the contribution is re-described as rarity weighting of sparse codes if the random dictionary is within noise on at least 8.
The first rule holds: \method is significantly ahead of the random dictionary with idf on 16 of 17 bootstrapped comparisons (the exception is XLCoST C\#, a tie), by 0.10--0.55 MRR on formal mathematics and 0.07--0.30 on knowledge-graph and cross-lingual terms, and the random dictionary with idf is below the plain dense state on every row. The biomedical tasks, run the same way after they joined the main table, agree: \method leads the random dictionary with idf by 0.10--0.13 MRR on both Bio-ML splits and on Anatomy ($p<0.001$), and the random dictionary stays below dense there too.
Sparsity plus rarity weighting is therefore not enough; the learned features carry the result, and the weighting is what makes them usable.
The calibration matched the mean number of active features (122.6 vs.\ 127.8) but not its distribution: with one global threshold on anisotropic activations the random code's median is 50 features and 11\% of its features never fire on the calibration sample, whereas the SAE's thresholds are per feature.

\paragraph{idf from a held-out library.}
The idf of the main table is computed over the candidate pool.
Computing it instead from 3,000 theorems sampled from set.mm (for Metamath) and 3,000 library terms from the HOL Light distribution (for HOL Light), rendered as the test statements are and encoded with the same model and SAE, gives MRR 0.588 on Lean~4 $\to$ Metamath (pool idf 0.661; dense 0.210) and 0.906 on $\to$ HOL Light (0.898).
By the pre-registered threshold (a drop of more than 0.05) the Metamath result makes the choice of reference collection a stated limitation; a library idf is nonetheless far above every dense baseline on both tasks.

\paragraph{A random sample of failures.}
Appendix~\ref{app:cases} selects case studies by rule; here five failures per task on the same ten tasks (gold not ranked first by \method) were drawn uniformly at random (seed 0) and decomposed in the same way, into the features the query shares with the gold counterpart and with the wrong top-1 candidate.
Failure rates range from 0.5\% of XLCoST queries and 4\% of NELL $\to$ DBpedia to 46\% on Lean~4 $\to$ Metamath and 34--37\% on MultiFarm; when \method fails, the gold is still within the top five in 57--100\% of cases, and Dense ranks it first in only 2--10\% of our failures on Metamath, HOL Light, MultiFarm Arabic and Valentine, 16--33\% on MultiFarm Chinese and Russian, YAGO and XLCoST, and 45\% and 60\% on Isabelle and NELL $\to$ DBpedia, so most failures are shared by both representations.
The features carrying the wrong matches have the same explanation-type mix as those carrying the gold (about 62\% \emph{concept}, 16\% \emph{notation}, 5\% \emph{generic} on both sides); the single largest contributor across all sampled failures is a \emph{generic} feature (``specific structured information or categories'') on XLCoST.
The 316 distinct features involved are released with their Neuronpedia explanations for a human check of the automatic names.

\section{Statistical significance}
\label{app:significance}
\begin{table*}[tp]
  \centering\scriptsize
  \setlength{\tabcolsep}{3.5pt}
  \begin{tabular}{@{}lrllllll@{}}
\toprule
 & & \multicolumn{2}{c}{vs.\ Dense} & \multicolumn{2}{c}{vs.\ Dense$_{-\mathrm{PC1}}$} & \multicolumn{2}{c}{vs.\ Strings} \\
\cmidrule(lr){3-4}\cmidrule(lr){5-6}\cmidrule(lr){7-8}
Task & $n$ & $\Delta$ [95\% CI] & $p$ & $\Delta$ [95\% CI] & $p$ & $\Delta$ [95\% CI] & $p$ \\
\midrule
Lean 4 $\to$ Isabelle & 483 & +0.048 [+0.027, +0.071] & $<$.001 & +0.034 [+0.014, +0.055] & $<$.001 & +0.136 [+0.105, +0.168] & $<$.001 \\
Lean 4 $\to$ Metamath & 482 & +0.451 [+0.418, +0.485] & $<$.001 & +0.510 [+0.475, +0.546] & $<$.001 & +0.046 [+0.002, +0.090] & 0.02 \\
Lean 4 $\to$ HOL Light & 317 & +0.359 [+0.315, +0.405] & $<$.001 & +0.367 [+0.322, +0.412] & $<$.001 & +0.157 [+0.114, +0.200] & $<$.001 \\
XLCoST Java & 3,918 & +0.014 [+0.013, +0.016] & $<$.001 & +0.015 [+0.013, +0.016] & $<$.001 & -0.005 [-0.006, -0.004] & 1.00 \\
XLCoST C++ & 3,895 & +0.008 [+0.007, +0.010] & $<$.001 & +0.009 [+0.008, +0.011] & $<$.001 & -0.008 [-0.009, -0.006] & 1.00 \\
XLCoST Python & 3,853 & +0.011 [+0.009, +0.012] & $<$.001 & +0.017 [+0.016, +0.019] & $<$.001 & +0.000 [-0.001, +0.002] & 0.25 \\
XLCoST C\# & 3,898 & +0.013 [+0.012, +0.015] & $<$.001 & +0.013 [+0.012, +0.015] & $<$.001 & -0.006 [-0.006, -0.005] & 1.00 \\
XLCoST JavaScript & 3,851 & +0.006 [+0.005, +0.008] & $<$.001 & +0.010 [+0.009, +0.012] & $<$.001 & -0.005 [-0.006, -0.004] & 1.00 \\
XLCoST PHP & 1,566 & +0.010 [+0.008, +0.011] & $<$.001 & +0.010 [+0.008, +0.011] & $<$.001 & +0.001 [-0.000, +0.003] & 0.07 \\
XLCoST C & 283 & +0.013 [+0.009, +0.016] & $<$.001 & +0.010 [+0.008, +0.013] & $<$.001 & -0.004 [-0.008, +0.000] & 0.96 \\
Valentine (F1) & 551 & +0.081 [+0.074, +0.089] & $<$.001 & +0.080 [+0.072, +0.089] & $<$.001 & -- \\
Valentine (MRR) & 551 & +0.090 [+0.075, +0.105] & $<$.001 & +0.095 [+0.082, +0.109] & $<$.001 & -- \\
NELL $\to$ DBpedia & 129 & +0.045 [+0.011, +0.081] & $<$.01 & +0.018 [-0.009, +0.048] & 0.11 & +0.040 [+0.014, +0.071] & $<$.001 \\
YAGO $\to$ Wikidata & 304 & +0.102 [+0.070, +0.136] & $<$.001 & +0.030 [+0.007, +0.054] & $<$.01 & +0.062 [+0.039, +0.087] & $<$.001 \\
MultiFarm zh $\to$ en & 246 & +0.104 [+0.058, +0.151] & $<$.001 & -0.006 [-0.040, +0.028] & 0.64 & +0.524 [+0.467, +0.578] & $<$.001 \\
MultiFarm ru $\to$ en & 246 & +0.121 [+0.079, +0.162] & $<$.001 & +0.014 [-0.017, +0.045] & 0.19 & +0.531 [+0.480, +0.581] & $<$.001 \\
MultiFarm ar $\to$ en & 239 & +0.127 [+0.088, +0.168] & $<$.001 & +0.019 [-0.016, +0.054] & 0.14 & +0.484 [+0.429, +0.539] & $<$.001 \\
Bio-ML valid & 634 & +0.117 [+0.094, +0.141] & $<$.001 & +0.086 [+0.064, +0.107] & $<$.001 & +0.038 [+0.017, +0.059] & $<$.001 \\
Bio-ML train & 3,845 & +0.080 [+0.071, +0.089] & $<$.001 & +0.057 [+0.049, +0.066] & $<$.001 & +0.026 [+0.019, +0.034] & $<$.001 \\
Anatomy & 1,516 & +0.074 [+0.061, +0.087] & $<$.001 & +0.021 [+0.009, +0.033] & $<$.001 & +0.152 [+0.137, +0.167] & $<$.001 \\
\bottomrule
\end{tabular}

  \caption{Paired bootstrap over queries (10,000 resamples): mean difference \method minus baseline, 95\% interval, and one-sided bootstrap $p$-value. Valentine has no string baseline.}
  \label{tab:significance}
\end{table*}
\begin{table*}[tp]
  \centering\scriptsize
  \setlength{\tabcolsep}{3.5pt}
  \begin{tabular}{@{}lrllllll@{}}
\toprule
 & & \multicolumn{2}{c}{vs.\ Dense} & \multicolumn{2}{c}{vs.\ Dense$_{-\mathrm{PC1}}$} & \multicolumn{2}{c}{vs.\ Strings} \\
\cmidrule(lr){3-4}\cmidrule(lr){5-6}\cmidrule(lr){7-8}
Task & $n$ & $\Delta$ [95\% CI] & $p$ & $\Delta$ [95\% CI] & $p$ & $\Delta$ [95\% CI] & $p$ \\
\midrule
Lean 4 $\to$ Isabelle & 483 & +0.034 [+0.018, +0.052] & $<$.001 & +0.032 [+0.015, +0.049] & $<$.001 & +0.152 [+0.120, +0.184] & $<$.001 \\
Lean 4 $\to$ Metamath & 482 & +0.141 [+0.109, +0.173] & $<$.001 & +0.150 [+0.118, +0.182] & $<$.001 & +0.172 [+0.132, +0.213] & $<$.001 \\
Lean 4 $\to$ HOL Light & 317 & +0.303 [+0.260, +0.348] & $<$.001 & +0.207 [+0.170, +0.245] & $<$.001 & +0.166 [+0.123, +0.208] & $<$.001 \\
XLCoST Java & 3,918 & +0.007 [+0.004, +0.009] & $<$.001 & +0.010 [+0.008, +0.012] & $<$.001 & -0.015 [-0.017, -0.014] & 1.00 \\
XLCoST C++ & 3,895 & -0.058 [-0.062, -0.053] & 1.00 & -0.053 [-0.058, -0.049] & 1.00 & -0.067 [-0.071, -0.063] & 1.00 \\
XLCoST Python & 3,853 & -0.011 [-0.013, -0.009] & 1.00 & -0.002 [-0.005, +0.000] & 0.96 & -0.024 [-0.027, -0.021] & 1.00 \\
XLCoST C\# & 3,898 & +0.005 [+0.003, +0.008] & $<$.001 & +0.010 [+0.008, +0.012] & $<$.001 & -0.017 [-0.019, -0.016] & 1.00 \\
XLCoST JavaScript & 3,851 & -0.004 [-0.005, -0.002] & 1.00 & +0.001 [-0.001, +0.002] & 0.19 & -0.017 [-0.019, -0.016] & 1.00 \\
XLCoST PHP & 1,566 & +0.003 [+0.001, +0.005] & $<$.01 & +0.006 [+0.003, +0.008] & $<$.001 & -0.013 [-0.015, -0.011] & 1.00 \\
XLCoST C & 283 & +0.000 [-0.002, +0.002] & 0.48 & -0.004 [-0.007, -0.002] & 1.00 & -0.025 [-0.029, -0.020] & 1.00 \\
Valentine (F1) & 551 & +0.037 [+0.030, +0.044] & $<$.001 & +0.046 [+0.039, +0.053] & $<$.001 & -- \\
Valentine (MRR) & 551 & +0.075 [+0.067, +0.083] & $<$.001 & +0.051 [+0.042, +0.059] & $<$.001 & -- \\
NELL $\to$ DBpedia & 129 & +0.009 [-0.021, +0.040] & 0.27 & -0.002 [-0.028, +0.024] & 0.57 & +0.022 [-0.004, +0.052] & 0.05 \\
YAGO $\to$ Wikidata & 304 & +0.056 [+0.030, +0.083] & $<$.001 & +0.000 [-0.021, +0.022] & 0.49 & +0.045 [+0.024, +0.067] & $<$.001 \\
MultiFarm zh $\to$ en & 246 & +0.128 [+0.087, +0.170] & $<$.001 & +0.008 [-0.022, +0.038] & 0.31 & +0.557 [+0.508, +0.606] & $<$.001 \\
MultiFarm ru $\to$ en & 246 & +0.065 [+0.022, +0.107] & $<$.01 & -0.024 [-0.059, +0.010] & 0.92 & +0.492 [+0.441, +0.545] & $<$.001 \\
MultiFarm ar $\to$ en & 239 & +0.094 [+0.050, +0.139] & $<$.001 & +0.002 [-0.038, +0.044] & 0.46 & +0.434 [+0.379, +0.488] & $<$.001 \\
Bio-ML valid & 634 & +0.075 [+0.052, +0.099] & $<$.001 & +0.069 [+0.047, +0.091] & $<$.001 & -0.000 [-0.021, +0.020] & 0.51 \\
Bio-ML train & 3,845 & +0.062 [+0.053, +0.071] & $<$.001 & +0.046 [+0.038, +0.055] & $<$.001 & -0.000 [-0.008, +0.007] & 0.51 \\
Anatomy & 1,516 & +0.043 [+0.029, +0.058] & $<$.001 & -0.027 [-0.040, -0.015] & 1.00 & +0.110 [+0.094, +0.126] & $<$.001 \\
\bottomrule
\end{tabular}

  \caption{The same paired bootstrap for the Llama-3.1-8B runs (layer 15, Llama Scope 131k).}
  \label{tab:significancellama}
\end{table*}

For every comparison in the main table we take the per-query quantity that the metric averages (the official per-query MRR for miniF2F and XLCoST, the per-pair F1 or MRR for Valentine, the reciprocal rank of the gold counterpart for the ontology and knowledge-graph tasks), resample the queries with replacement 10,000 times, and report the mean difference \method minus baseline with its 95\% percentile interval and the one-sided bootstrap $p$-value $P(\Delta\le 0)$ (Table~\ref{tab:significance}).
Every gain over Dense has $p<0.004$.
The gain over \pcone has $p<0.01$ on every row except the three MultiFarm pairs and NELL $\to$ DBpedia, where the interval covers zero.
Against strings, \method is significantly ahead on all formal-mathematics, knowledge-graph and ontology tasks (Lean~4 $\to$ Metamath marginally, $p=0.02$), tied on XLCoST Python, PHP and C, and significantly behind word TF-IDF on XLCoST Java, C++, C\# and JavaScript.
Table~\ref{tab:significanceextras} applies the same bootstrap to the post-hoc controls of Appendix~\ref{app:controls}.
Table~\ref{tab:significancellama} repeats the analysis for the Llama-3.1-8B runs: the gains over dense stay significant on the formal-mathematics, Valentine, YAGO, MultiFarm and biomedical tasks, while NELL $\to$ DBpedia is a tie, XLCoST is mixed (small significant gains on Java, C\# and PHP, deficits on C++, Python and JavaScript), and \pcone is significantly ahead on Anatomy.

\begin{table*}[tp]
  \centering\scriptsize\setlength{\tabcolsep}{3.5pt}
  \begin{tabular}{@{}lllrrrr@{}}
\toprule
Row & Comparison & $n$ & \method & other & $\Delta$ [95\% CI] & $p$ \\
\midrule
\multicolumn{7}{@{}l}{\emph{S1: random dictionary of the same width and sparsity}} \\
Lean 4 $\to$ Isabelle & Random + idf & 483 & 0.961 & 0.865 & +0.096 [+0.068, +0.125] & $<$0.001 \\
Lean 4 $\to$ Isabelle & Random, unw. & 483 & 0.961 & 0.845 & +0.116 [+0.088, +0.145] & $<$0.001 \\
Lean 4 $\to$ Metamath & Random + idf & 482 & 0.661 & 0.146 & +0.516 [+0.476, +0.555] & $<$0.001 \\
Lean 4 $\to$ Metamath & Random, unw. & 482 & 0.661 & 0.156 & +0.505 [+0.466, +0.543] & $<$0.001 \\
Lean 4 $\to$ HOL Light & Random + idf & 317 & 0.898 & 0.349 & +0.549 [+0.500, +0.596] & $<$0.001 \\
Lean 4 $\to$ HOL Light & Random, unw. & 317 & 0.898 & 0.484 & +0.414 [+0.367, +0.461] & $<$0.001 \\
XLCoST Java & Random + idf & 3918 & 0.480 & 0.420 & +0.060 [+0.056, +0.063] & $<$0.001 \\
XLCoST Java & Random, unw. & 3918 & 0.480 & 0.467 & +0.013 [+0.011, +0.016] & $<$0.001 \\
XLCoST Cpp & Random + idf & 3895 & 0.478 & 0.317 & +0.161 [+0.155, +0.167] & $<$0.001 \\
XLCoST Cpp & Random, unw. & 3895 & 0.478 & 0.387 & +0.091 [+0.087, +0.096] & $<$0.001 \\
XLCoST Python & Random + idf & 3853 & 0.485 & 0.471 & +0.014 [+0.013, +0.016] & $<$0.001 \\
XLCoST Python & Random, unw. & 3853 & 0.485 & 0.472 & +0.013 [+0.011, +0.015] & $<$0.001 \\
XLCoST Csharp & Random + idf & 3898 & 0.479 & 0.479 & +0.000 [-0.002, +0.002] & 0.366 \\
XLCoST Csharp & Random, unw. & 3898 & 0.479 & 0.483 & -0.004 [-0.006, -0.002] & 1.000 \\
XLCoST Javascript & Random + idf & 3851 & 0.482 & 0.470 & +0.012 [+0.009, +0.014] & $<$0.001 \\
XLCoST Javascript & Random, unw. & 3851 & 0.482 & 0.478 & +0.003 [+0.001, +0.005] & $<$0.001 \\
XLCoST PHP & Random + idf & 1566 & 0.442 & 0.426 & +0.016 [+0.013, +0.019] & $<$0.001 \\
XLCoST PHP & Random, unw. & 1566 & 0.442 & 0.421 & +0.021 [+0.017, +0.024] & $<$0.001 \\
XLCoST C & Random + idf & 283 & 0.405 & 0.336 & +0.069 [+0.059, +0.079] & $<$0.001 \\
XLCoST C & Random, unw. & 283 & 0.405 & 0.295 & +0.110 [+0.099, +0.122] & $<$0.001 \\
NELL $\to$ DBpedia & Random + idf & 129 & 0.976 & 0.901 & +0.075 [+0.032, +0.121] & $<$0.001 \\
NELL $\to$ DBpedia & Random, unw. & 129 & 0.976 & 0.901 & +0.075 [+0.032, +0.123] & $<$0.001 \\
YAGO $\to$ Wikidata & Random + idf & 304 & 0.962 & 0.797 & +0.165 [+0.126, +0.204] & $<$0.001 \\
YAGO $\to$ Wikidata & Random, unw. & 304 & 0.962 & 0.797 & +0.165 [+0.126, +0.206] & $<$0.001 \\
MultiFarm zh $\to$ en & Random + idf & 246 & 0.754 & 0.533 & +0.221 [+0.165, +0.274] & $<$0.001 \\
MultiFarm zh $\to$ en & Random, unw. & 246 & 0.754 & 0.482 & +0.272 [+0.218, +0.327] & $<$0.001 \\
MultiFarm ru $\to$ en & Random + idf & 246 & 0.765 & 0.542 & +0.222 [+0.170, +0.275] & $<$0.001 \\
MultiFarm ru $\to$ en & Random, unw. & 246 & 0.765 & 0.448 & +0.316 [+0.261, +0.370] & $<$0.001 \\
MultiFarm ar $\to$ en & Random + idf & 239 & 0.722 & 0.425 & +0.297 [+0.245, +0.351] & $<$0.001 \\
MultiFarm ar $\to$ en & Random, unw. & 239 & 0.722 & 0.368 & +0.354 [+0.301, +0.410] & $<$0.001 \\
Bio-ML valid & Random + idf & 634 & 0.817 & 0.692 & +0.126 [+0.100, +0.152] & $<$0.001 \\
Bio-ML valid & Random, unw. & 634 & 0.817 & 0.680 & +0.137 [+0.111, +0.162] & $<$0.001 \\
Bio-ML train & Random + idf & 3845 & 0.815 & 0.711 & +0.105 [+0.095, +0.115] & $<$0.001 \\
Bio-ML train & Random, unw. & 3845 & 0.815 & 0.693 & +0.123 [+0.113, +0.133] & $<$0.001 \\
Anatomy & Random + idf & 1516 & 0.886 & 0.788 & +0.098 [+0.083, +0.114] & $<$0.001 \\
Anatomy & Random, unw. & 1516 & 0.886 & 0.791 & +0.094 [+0.078, +0.111] & $<$0.001 \\
Valentine (F1) & Random + idf & 551 & 0.614 & 0.482 & +0.132 [+0.118, +0.147] & $<$0.001 \\
Valentine (F1) & Random, unw. & 551 & 0.614 & 0.460 & +0.154 [+0.137, +0.172] & $<$0.001 \\
Valentine (MRR) & Random + idf & 551 & 0.856 & 0.647 & +0.209 [+0.187, +0.231] & $<$0.001 \\
Valentine (MRR) & Random, unw. & 551 & 0.856 & 0.654 & +0.202 [+0.179, +0.226] & $<$0.001 \\
\bottomrule
  \end{tabular}
  \caption{Paired bootstrap for the post-hoc controls of Appendix~\ref{app:controls}: \method minus the comparison, per row (XLCoST per language), with the 95\% interval and one-sided $p$.}\label{tab:significanceextras}
\end{table*}

\begin{table*}[tp]
  \centering\scriptsize\setlength{\tabcolsep}{3.5pt}
  \begin{tabular}{@{}lllrrrr@{}}
\toprule
Row & Comparison & $n$ & \method & other & $\Delta$ [95\% CI] & $p$ \\
\midrule
\multicolumn{7}{@{}l}{\emph{S2: stronger dense controls}} \\
Lean 4 $\to$ Isabelle & $-$36 PCs & 483 & 0.961 & 0.928 & +0.033 [+0.014, +0.052] & $<$0.001 \\
Lean 4 $\to$ Isabelle & z-score & 483 & 0.961 & 0.942 & +0.019 [+0.001, +0.038] & 0.020 \\
Lean 4 $\to$ Isabelle & ZCA & 483 & 0.961 & 0.946 & +0.015 [-0.002, +0.032] & 0.041 \\
Lean 4 $\to$ Isabelle & dense-idf & 483 & 0.961 & 0.942 & +0.019 [+0.001, +0.037] & 0.020 \\
Lean 4 $\to$ Isabelle & sparse dense-idf & 483 & 0.961 & 0.914 & +0.047 [+0.026, +0.069] & $<$0.001 \\
Lean 4 $\to$ Metamath & $-$36 PCs & 482 & 0.661 & 0.401 & +0.260 [+0.225, +0.297] & $<$0.001 \\
Lean 4 $\to$ Metamath & z-score & 482 & 0.661 & 0.257 & +0.404 [+0.369, +0.439] & $<$0.001 \\
Lean 4 $\to$ Metamath & ZCA & 482 & 0.661 & 0.465 & +0.196 [+0.163, +0.230] & $<$0.001 \\
Lean 4 $\to$ Metamath & dense-idf & 482 & 0.661 & 0.251 & +0.410 [+0.375, +0.446] & $<$0.001 \\
Lean 4 $\to$ Metamath & sparse dense-idf & 482 & 0.661 & 0.235 & +0.426 [+0.392, +0.461] & $<$0.001 \\
Lean 4 $\to$ HOL Light & $-$36 PCs & 317 & 0.898 & 0.728 & +0.170 [+0.133, +0.208] & $<$0.001 \\
Lean 4 $\to$ HOL Light & z-score & 317 & 0.898 & 0.707 & +0.191 [+0.154, +0.230] & $<$0.001 \\
Lean 4 $\to$ HOL Light & ZCA & 317 & 0.898 & 0.786 & +0.112 [+0.080, +0.145] & $<$0.001 \\
Lean 4 $\to$ HOL Light & dense-idf & 317 & 0.898 & 0.706 & +0.192 [+0.154, +0.231] & $<$0.001 \\
Lean 4 $\to$ HOL Light & sparse dense-idf & 317 & 0.898 & 0.605 & +0.293 [+0.250, +0.335] & $<$0.001 \\
XLCoST Java & $-$36 PCs & 3918 & 0.994 & 0.489 & +0.505 [+0.495, +0.514] & $<$0.001 \\
XLCoST Java & z-score & 3918 & 0.994 & 0.460 & +0.533 [+0.523, +0.543] & $<$0.001 \\
XLCoST Java & ZCA & 3918 & 0.994 & 0.463 & +0.531 [+0.521, +0.541] & $<$0.001 \\
XLCoST Java & dense-idf & 3918 & 0.994 & 0.460 & +0.533 [+0.523, +0.543] & $<$0.001 \\
XLCoST Java & sparse dense-idf & 3918 & 0.994 & 0.466 & +0.527 [+0.517, +0.537] & $<$0.001 \\
XLCoST Cpp & $-$36 PCs & 3895 & 0.985 & 0.483 & +0.503 [+0.493, +0.512] & $<$0.001 \\
XLCoST Cpp & z-score & 3895 & 0.985 & 0.477 & +0.508 [+0.498, +0.518] & $<$0.001 \\
XLCoST Cpp & ZCA & 3895 & 0.985 & 0.472 & +0.513 [+0.503, +0.524] & $<$0.001 \\
XLCoST Cpp & dense-idf & 3895 & 0.985 & 0.477 & +0.508 [+0.499, +0.518] & $<$0.001 \\
XLCoST Cpp & sparse dense-idf & 3895 & 0.985 & 0.475 & +0.510 [+0.500, +0.520] & $<$0.001 \\
XLCoST Python & $-$36 PCs & 3853 & 0.991 & 0.483 & +0.508 [+0.498, +0.517] & $<$0.001 \\
XLCoST Python & z-score & 3853 & 0.991 & 0.477 & +0.514 [+0.504, +0.523] & $<$0.001 \\
XLCoST Python & ZCA & 3853 & 0.991 & 0.405 & +0.586 [+0.575, +0.597] & $<$0.001 \\
XLCoST Python & dense-idf & 3853 & 0.991 & 0.477 & +0.514 [+0.505, +0.524] & $<$0.001 \\
XLCoST Python & sparse dense-idf & 3853 & 0.991 & 0.442 & +0.549 [+0.539, +0.560] & $<$0.001 \\
XLCoST Csharp & $-$36 PCs & 3898 & 0.994 & 0.488 & +0.506 [+0.497, +0.516] & $<$0.001 \\
XLCoST Csharp & z-score & 3898 & 0.994 & 0.459 & +0.535 [+0.525, +0.545] & $<$0.001 \\
XLCoST Csharp & ZCA & 3898 & 0.994 & 0.462 & +0.532 [+0.522, +0.542] & $<$0.001 \\
XLCoST Csharp & dense-idf & 3898 & 0.994 & 0.459 & +0.535 [+0.525, +0.545] & $<$0.001 \\
XLCoST Csharp & sparse dense-idf & 3898 & 0.994 & 0.464 & +0.530 [+0.520, +0.540] & $<$0.001 \\
XLCoST Javascript & $-$36 PCs & 3851 & 0.990 & 0.488 & +0.502 [+0.493, +0.512] & $<$0.001 \\
XLCoST Javascript & z-score & 3851 & 0.990 & 0.479 & +0.511 [+0.502, +0.521] & $<$0.001 \\
XLCoST Javascript & ZCA & 3851 & 0.990 & 0.481 & +0.510 [+0.500, +0.519] & $<$0.001 \\
XLCoST Javascript & dense-idf & 3851 & 0.990 & 0.479 & +0.511 [+0.502, +0.521] & $<$0.001 \\
XLCoST Javascript & sparse dense-idf & 3851 & 0.990 & 0.478 & +0.513 [+0.503, +0.522] & $<$0.001 \\
XLCoST PHP & $-$36 PCs & 1566 & 0.988 & 0.444 & +0.544 [+0.529, +0.558] & $<$0.001 \\
XLCoST PHP & z-score & 1566 & 0.988 & 0.437 & +0.551 [+0.536, +0.565] & $<$0.001 \\
XLCoST PHP & ZCA & 1566 & 0.988 & 0.388 & +0.600 [+0.583, +0.617] & $<$0.001 \\
\bottomrule
  \end{tabular}
  \addtocounter{table}{-1}\caption{Paired bootstrap for the post-hoc controls of Appendix~\ref{app:controls} (continued).}
\end{table*}

\begin{table*}[tp]
  \centering\scriptsize\setlength{\tabcolsep}{3.5pt}
  \begin{tabular}{@{}lllrrrr@{}}
\toprule
Row & Comparison & $n$ & \method & other & $\Delta$ [95\% CI] & $p$ \\
\midrule
\multicolumn{7}{@{}l}{\emph{S2: stronger dense controls, continued}} \\
XLCoST PHP & dense-idf & 1566 & 0.988 & 0.437 & +0.551 [+0.536, +0.566] & $<$0.001 \\
XLCoST PHP & sparse dense-idf & 1566 & 0.988 & 0.435 & +0.553 [+0.538, +0.568] & $<$0.001 \\
XLCoST C & $-$36 PCs & 283 & 0.979 & 0.316 & +0.663 [+0.625, +0.700] & $<$0.001 \\
XLCoST C & z-score & 283 & 0.979 & 0.393 & +0.585 [+0.549, +0.620] & $<$0.001 \\
XLCoST C & ZCA & 283 & 0.979 & 0.283 & +0.696 [+0.655, +0.738] & $<$0.001 \\
XLCoST C & dense-idf & 283 & 0.979 & 0.394 & +0.585 [+0.549, +0.620] & $<$0.001 \\
XLCoST C & sparse dense-idf & 283 & 0.979 & 0.397 & +0.581 [+0.546, +0.616] & $<$0.001 \\
NELL $\to$ DBpedia & $-$36 PCs & 129 & 0.976 & 0.973 & +0.003 [-0.016, +0.019] & 0.437 \\
NELL $\to$ DBpedia & z-score & 129 & 0.976 & 0.923 & +0.053 [+0.015, +0.094] & 0.003 \\
NELL $\to$ DBpedia & ZCA & 129 & 0.976 & 0.989 & -0.013 [-0.033, +0.003] & 0.952 \\
NELL $\to$ DBpedia & dense-idf & 129 & 0.976 & 0.923 & +0.053 [+0.016, +0.094] & 0.002 \\
NELL $\to$ DBpedia & sparse dense-idf & 129 & 0.976 & 0.925 & +0.051 [+0.008, +0.097] & 0.009 \\
YAGO $\to$ Wikidata & $-$36 PCs & 304 & 0.962 & 0.957 & +0.005 [-0.013, +0.023] & 0.283 \\
YAGO $\to$ Wikidata & z-score & 304 & 0.962 & 0.814 & +0.148 [+0.110, +0.188] & $<$0.001 \\
YAGO $\to$ Wikidata & ZCA & 304 & 0.962 & 0.964 & -0.003 [-0.018, +0.013] & 0.631 \\
YAGO $\to$ Wikidata & dense-idf & 304 & 0.962 & 0.812 & +0.150 [+0.111, +0.190] & $<$0.001 \\
YAGO $\to$ Wikidata & sparse dense-idf & 304 & 0.962 & 0.847 & +0.115 [+0.081, +0.151] & $<$0.001 \\
MultiFarm zh $\to$ en & $-$36 PCs & 246 & 0.754 & 0.250 & +0.504 [+0.451, +0.557] & $<$0.001 \\
MultiFarm zh $\to$ en & z-score & 246 & 0.754 & 0.714 & +0.040 [+0.002, +0.077] & 0.020 \\
MultiFarm zh $\to$ en & ZCA & 246 & 0.754 & 0.136 & +0.618 [+0.564, +0.669] & $<$0.001 \\
MultiFarm zh $\to$ en & dense-idf & 246 & 0.754 & 0.710 & +0.044 [+0.006, +0.081] & 0.012 \\
MultiFarm zh $\to$ en & sparse dense-idf & 246 & 0.754 & 0.582 & +0.172 [+0.124, +0.221] & $<$0.001 \\
MultiFarm ru $\to$ en & $-$36 PCs & 246 & 0.765 & 0.259 & +0.506 [+0.449, +0.561] & $<$0.001 \\
MultiFarm ru $\to$ en & z-score & 246 & 0.765 & 0.713 & +0.052 [+0.013, +0.091] & 0.004 \\
MultiFarm ru $\to$ en & ZCA & 246 & 0.765 & 0.156 & +0.609 [+0.555, +0.662] & $<$0.001 \\
MultiFarm ru $\to$ en & dense-idf & 246 & 0.765 & 0.710 & +0.055 [+0.015, +0.094] & 0.004 \\
MultiFarm ru $\to$ en & sparse dense-idf & 246 & 0.765 & 0.582 & +0.182 [+0.135, +0.229] & $<$0.001 \\
MultiFarm ar $\to$ en & $-$36 PCs & 239 & 0.722 & 0.240 & +0.482 [+0.428, +0.535] & $<$0.001 \\
MultiFarm ar $\to$ en & z-score & 239 & 0.722 & 0.651 & +0.071 [+0.036, +0.108] & $<$0.001 \\
MultiFarm ar $\to$ en & ZCA & 239 & 0.722 & 0.170 & +0.552 [+0.498, +0.606] & $<$0.001 \\
MultiFarm ar $\to$ en & dense-idf & 239 & 0.722 & 0.644 & +0.078 [+0.042, +0.116] & $<$0.001 \\
MultiFarm ar $\to$ en & sparse dense-idf & 239 & 0.722 & 0.545 & +0.177 [+0.130, +0.223] & $<$0.001 \\
Valentine (F1) & $-$36 PCs & 551 & 0.614 & 0.468 & +0.146 [+0.130, +0.161] & $<$0.001 \\
Valentine (F1) & z-score & 551 & 0.614 & 0.459 & +0.155 [+0.138, +0.171] & $<$0.001 \\
Valentine (F1) & ZCA & 551 & 0.614 & 0.411 & +0.204 [+0.186, +0.222] & $<$0.001 \\
Valentine (F1) & dense-idf & 551 & 0.614 & 0.399 & +0.215 [+0.196, +0.233] & $<$0.001 \\
Valentine (F1) & sparse dense-idf & 551 & 0.614 & 0.208 & +0.406 [+0.385, +0.427] & $<$0.001 \\
Valentine (MRR) & $-$36 PCs & 551 & 0.856 & 0.548 & +0.308 [+0.287, +0.329] & $<$0.001 \\
Valentine (MRR) & z-score & 551 & 0.856 & 0.529 & +0.327 [+0.305, +0.349] & $<$0.001 \\
Valentine (MRR) & ZCA & 551 & 0.856 & 0.503 & +0.353 [+0.331, +0.373] & $<$0.001 \\
Valentine (MRR) & dense-idf & 551 & 0.856 & 0.485 & +0.371 [+0.350, +0.393] & $<$0.001 \\
Valentine (MRR) & sparse dense-idf & 551 & 0.856 & 0.390 & +0.466 [+0.445, +0.485] & $<$0.001 \\
\bottomrule
  \end{tabular}
  \addtocounter{table}{-1}\caption{Paired bootstrap for the post-hoc controls of Appendix~\ref{app:controls} (continued).}
\end{table*}

\section{What did not work}
\label{app:tacit}

Generated contexts (Appendix~\ref{app:generated}) and a drug-name task with shared metadata, where the shared text leaked the answer to both sides.
Pooling hidden states before SAE encoding, which drops features that fire on part of a symbol.
A ``universal'' idf computed from Pile densities instead of the matched collection, which was worse on every task (0.190 vs.\ 0.661 MRR on Metamath) because it does not know which features are generic \emph{for these two systems}.
Qwen2.5-7B-Instruct as a context generator (broken fragments and factual errors).
Using synonyms as context, which hands over the other side's label.
Smaller candidate pools (a 59-pair Valentine subset gave the opposite verdict to the full 551 pairs), which is why every reported number is on full data.

\end{document}